%% file: manuscript.tex
\documentclass[lettersize,journal]{IEEEtran}

\input{preamble}

\input{notation}

\begin{document}

\title{Generative Graphical Inverse Kinematics}

\author{%
  Oliver Limoyo{$^{\dagger, a}$},
  Filip Mari\'c{$^{\dagger, a, b}$},
  Matthew Giamou{$^{a}$},
  Petra Alexson{$^{a}$},
  Ivan Petrovi\'c{$^{a, b}$},
  Jonathan Kelly{$^{a}$}
  \thanks{$^\dagger$Denotes equal contribution.}
  \thanks{$^a$Filip Mari\'c, Oliver Limoyo, Matthew Giamou, Petra Alexson, Ivan Petrovi\'c and Jonathan Kelly are with the Space and Terrestrial Autonomous Robotic Systems Laboratory, University of Toronto, Institute for Aerospace Studies, Toronto, Canada. \{\texttt{<first name>.<last name>@robotics.utias.utoronto.ca}\}}
  \thanks{$^b$Filip Mari\'c and Ivan Petrovi\'c are with the Laboratory for Autonomous Systems and Mobile Robotics, University of Zagreb, Faculty of Electrical Engineering and Computing, Zagreb, Croatia. \{\texttt{<first name>.<last name>@fer.hr}\}}
}

\maketitle

\begin{abstract}
Quickly and reliably finding accurate inverse kinematics (IK) solutions remains a challenging problem for many robot manipulators.
 Existing numerical solvers are broadly applicable but typically only produce a single solution and rely on local search techniques to minimize nonconvex objective functions.
More recent learning-based approaches that approximate the entire feasible set of solutions have shown promise as a means to generate multiple fast and accurate IK results in parallel.
 However, existing learning-based techniques have a significant drawback: each robot of interest requires a specialized model that must be trained from scratch.
To address this key shortcoming, we propose a novel distance-geometric robot representation coupled with a graph structure that allows us to leverage the sample efficiency of Euclidean equivariant functions and the generalizability of graph neural networks (GNNs).
Our approach is generative graphical inverse kinematics (GGIK), the first learned IK solver able to accurately and efficiently produce a large number of diverse solutions in parallel while also displaying the ability to generalize---a single learned model can be used to produce IK solutions for a variety of different robots.
When compared to several other learned IK methods, GGIK provides more accurate solutions with the same amount of data.
GGIK can generalize reasonably well to robot manipulators \emph{unseen} during training.
Additionally, GGIK can learn a constrained distribution that encodes joint limits and scales efficiently to larger robots and a high number of sampled solutions.
Finally, GGIK can be used to complement local IK solvers by providing reliable initializations for a local optimization process.
\end{abstract}

\section{Introduction}
\label{sec:intro}

Robotic manipulation tasks are naturally defined in terms of end-effector poses. 
However, the configuration of a manipulator is typically specified in terms of joint angles, and determining the joint configuration(s) that correspond to a given end-effector pose requires solving the \emph{inverse kinematics} (IK) problem.
For redundant manipulators (i.e., those with more than six degrees of freedom or DOF), target poses may be reachable by an infinite set of feasible configurations.
While redundancy allows high-level algorithms (e.g., motion planners) to choose configurations that best fit the overall task, it makes solving IK substantially more involved.

Since the full set of IK solutions cannot, in general, be derived analytically for redundant manipulators, individual configurations reaching a target pose are found by locally searching the configuration space using numerical optimization methods and geometric heuristics.
The search space is sometimes reduced by constraining solutions to have particular properties (e.g., collision avoidance, manipulability).
While existing numerical solvers are broadly applicable, they usually only produce one single solution at a time and rely on incremental search techniques to minimize highly nonconvex objective functions.

These limitations have motivated the use of learned IK models that approximate the entire feasible set of solutions.
In terms of success rate, learned models that output individual solutions are able to compete with the best numerical IK solvers when high accuracy is not required~\cite{vonoehsen_comparison_2020}.
Data-driven methods are also useful for integrating abstract criteria such as ``human-like'' poses or motions~\cite{aristidouInverseKinematicsTechniques2018}.
Generative approaches~\cite{ren2020learning, ho2022selective} have demonstrated the ability to rapidly produce a large number of approximate IK solutions and to even model the entire feasible set for specific robots~\cite{ames2021ikflow}.
Access to a large number of configurations fitting desired constraints has proven beneficial in motion planning applications~\cite{lembono2021learning}.
Unfortunately, these learned models, parameterized by deep neural networks (DNNs), require specific configuration and end-effector input-output vector pairs for training (by design).
In turn, it is not possible to generalize learned solutions to robots that vary in link geometry and DOF.
Ultimately, this drawback limits the utility of learning for IK over well-established numerical methods that are easier to implement and to generalize~\cite{beeson2015trac}.
In this paper, we describe a generative graphical inverse kinematics (GGIK) model and explain its capacity to simultaneously represent general (i.e., not tied to a single robot manipulator model or geometry) IK mappings and to produce approximations of entire feasible sets of solutions.
In contrast to existing DNN-based approaches~\cite{ren2020learning, lembono2021learning, vonoehsen_comparison_2020, ho2022selective, ames2021ikflow}, we explore a new path towards learning generalized IK by adopting a \emph{graphical} description of robot kinematics.
This graph-based description, first introduced in~\cite{porta_branch-and-prune_2005} and extended and generalized in~\cite{2021_Maric_Riemannian_B}, is formulated using distances between points rigidly attached to the robot structure.
IK can then be cast as the problem of finding unknown distances (i.e., weights in a distance graph), allowing us to make use of graph neural networks (GNNs) to capture varying robot geometries and DOF within a single learned model.
Furthermore, the graphical formulation exposes the symmetry and Euclidean equivariance of the IK problem that stems from the spatial nature of robot manipulators.
We exploit this symmetry by encoding it into the structure of our model architecture to efficiently learn accurate IK solutions.
When compared to several other learned IK methods, GGIK provides more accurate results. 
GGIK is also able to generalize reasonably well to \emph{unseen} robot manipulators.
We show that GGIK can complement local IK solvers by providing reliable initializations.
Our code and models are open source.\footnote{\texttt{\url{https://github.com/utiasSTARS/generative-graphik}}}

\section{Related Work}
\label{sec:relatedwork}

Our work leverages prior research in several diverse areas, including classical IK methods, learning for IK and motion planning, and generative modelling on graphs.
We briefly summarize relevant literature in each of these areas below.

\subsection{Inverse Kinematics}
Robotic manipulators with six or fewer DOF can reach any feasible end-effector pose~\cite{lee1988new} with up to sixteen different configurations that can be determined analytically using various representations~\cite{manocha1994efficient, husty2007new}.
For example, the IKFast algorithm~\cite{diankov2010automated} uses symbolic computation to recover the full set of configurations as solutions to  polynomial equations that are robot-specific.
However, analytically solving IK for redundant manipulators is generally infeasible and numerical optimization methods or geometric heuristics must be used instead.
Simple instances of the IK problem involving small changes in the end-effector pose (e.g., for kinematic control or reactive planning) are also known as \emph{differential} IK.
These instances can be reliably solved by computing required configuration changes using a first-order Taylor series expansion of the manipulator's forward kinematics~\cite{whitney1969resolved,sciavicco1986coordinate}.
As an extension of this approach, the infinite solution space afforded by redundant DOF can be used to optimize secondary objectives, such as avoiding joint limits, for example, through redundancy resolution techniques~\cite{nakamura1987task}.

By incrementally ``guiding'' the end-effector along a path in the task space, the differential IK approach can also be used to reach end-effector poses far from the initial configuration (e.g., a pose enabling a desired grasp)~\cite{lynch2017modern}.
This typically amounts to solving a sequence of convex optimization problems~\cite{boyd2004convex} such as quadratic programs (QPs) using general-purpose solvers~\cite{goldfarb1983numerically, wachter2006implementation, bambade2022prox}.
However, these \emph{closed-loop IK} techniques are particularly vulnerable to singularities and may require a user-specified end-effector path that reaches the goal~\cite{siciliano2010robotics}.

The ``guidance-free'' or full IK problem \cite{manipulation} that appears in applications such as motion planning is usually solved with first-order~\cite{engell2009projected, beeson2015trac} or second-order~\cite{deo1993adaptive, erleben_solving_2019} nonlinear programming methods formulated over joint angles.
These methods have robust theoretical underpinnings~\cite{nocedal1999numerical} and are featured in popular libraries such as KDL~\cite{kdl-url} and TRAC-IK~\cite{beeson2015trac}, which (approximately) support a wide range of constraints through the addition of penalties to the cost function.
However, the highly nonconvex nature of the problem makes them susceptible to local minima, often requiring multiple initial guesses before returning a feasible configuration, if at all.
Such issues can sometimes be circumvented through the use of heuristics such as the cyclic coordinate descent (CCD) algorithm~\cite{kenwright2012inverse}, which iteratively adjusts joint angles using simple geometric expressions.
Similarly, the FABRIK~\cite{Aristidou_2011} algorithm solves the IK problem using iterative forward and backward passes over joint positions to quickly find solutions.

Some alternative IK solvers forgo the joint angle parametrization in favour of Cartesian coordinates and geometric representations~\cite{dejalonTwentyfiveYearsNatural2007}.
Dai et al.~\cite{dai_global_2019} use a Cartesian parameterization together with a piecewise-convex relaxation of $\LieGroupSO{3}$ to formulate IK as a mixed-integer linear program, while Yenamandra et al.~\cite{yenamandra_convex_2019} apply a similar relaxation to formulate IK as a semidefinite program.
Naour et al.~\cite{le2019kinematics} express IK as a nonlinear program over inter-point distances, showing that solutions can be recovered for unconstrained articulated bodies.
The distance-geometric robot model introduced in~\cite{porta_branch-and-prune_2005} was used in \cite{2021_Maric_Riemannian_B} and \cite{2022_Giamou_Convex} to produce Riemannian and convex optimization-based IK formulations, respectively.
Our learning architecture adopts this distance-geometric paradigm to construct a graphical representation of the IK problem that can be leveraged by a GNN, maintaining the generalization capability inherent to conventional approaches.
In contrast to conventional methods, however, our model outputs a distribution representing the IK solution set rather than individual solutions only.

\subsection{Learning Inverse Kinematics}

Jordan and Rumelhart show in~\cite{jordan1992forward} that the non-uniqueness of IK solutions presents a major difficulty for learning algorithms, which often yield erroneous models that ``average" the nonconvex feasible set.
However, Bullock et al.~\cite{bullock1993self} observe that relating end-effector and configuration \emph{directions} instead of positions (i.e., working in the velocity space) ensures that the space of feasible inputs can be linearly approximated at each point along a trajectory, ensuring that their superposition is also feasible.
%
B\'{o}csi et al.~\cite{bocsi2011learning} use an SVM to parameterize a quadratic program whose solutions match those of position-only IK for particular workspace regions.
In computer graphics, Villegas et al.~\cite{villegasNeuralKinematicNetworks2018} apply a recursive neural network model to solve a highly constrained IK instance for motion transfer between skeletons with different bone lengths.
In this work, we show that a GNN-based IK solver allows a higher degree of generalization by capturing not only different link lengths, but also different numbers of DOF.

Recently, generative models have demonstrated the potential to represent the full set of IK solutions.
A number of invertible architectures~\cite{ardizzone2018analyzing, kruse2021benchmarking} have been able to successfully learn the feasible set for 2D kinematic chains.
Generative adversarial networks (GANs) have been applied to learn the inverse kinematics and dynamics of an 8-DOF robot~\cite{ren2020learning}. 
The learned model is used to improve motion planning performance by sampling configurations constrained by link positions and (partial) orientations~\cite{lembono2021learning}.
Ho et al.~\cite{ho2022selective} have proposed a model that retrieves configurations reaching a target position by decoding \emph{posture indices} for the closest position from a database of positions.
%
%
A series of individual networks have been leveraged to sequentially generate distributions of joint values of a manipulator in an auto-regressive manner~\cite{2022_Bensadoun_Neural}. 
Finally, Ames et al.~\cite{ames2021ikflow} describe IKFlow, a model that relies on normalizing flows to generate a distribution of IK solutions for a desired end-effector pose.
Our architecture differs from previous work by allowing the learned distribution to generalize to a larger class of robots, conveniently removing the requirement of training an entirely new model from scratch for each specific robot.

\subsection{Learning for Motion Planning}
Prior applications of learning in motion planning include warm-starting optimization-based methods~\cite{ichnowskiDeepLearningGrasp}, learning distributions for sampling-based algorithms~\cite{khan2020graph, ichter2018learning}, directly learning a motion planner~\cite{qureshi2020motion}, and learning cost functions for joint grasp and motion optimization with diffusion models \cite{10161569}.
In \cite{ichter2018learning}, Ichter et al. introduce a learning-based generative model, based on the conditional variational autoencoder (CVAE), to help solve various 2D and 3D navigation planning problems.
The learned model is applied to sample the state space in a manner that is biased towards favourable regions (i.e., where a solution is more likely to be found).
GGIK shares the same premise, to learn an initialization or sampler that improves upon the uniform sampling that is traditionally performed for both sampling-based motion planning and ``assumption-free'' IK.
However, although our methodology is similar in some respects, IK introduces additional challenges from the perspective of learning.
Specifically, for the IK problem, we aim to capture multiple solutions and to handle multiple manipulator structures.

\subsection{Generative Models on Graphs}

GGIK is a deep generative model of graphs. 
We provide a short review of generative models for graph representations, emphasizing recent deep learning-based methods, and point readers to~\cite{GNNBook-ch11-liao} for a more extensive survey.
Existing learning-based approaches typically make use of VAEs \cite{kipf2016variational}, generative adversarial networks (GANs) \cite{de2018molgan}, or deep auto-regressive methods \cite{li2018learning}. 
The authors of \cite{kipf2016variational} demonstrate how to extend the VAE framework to graph data.
Since GGIK uses a graph-based CVAE, it builds upon the theoretical groundwork of the VGAE (variational graph autoencoder) family of models from \cite{kipf2016variational}.
We use a conditional variant of the VGAE and a more expressive node decoder that is parametrized with a graph neural network as opposed to a simple inner product edge decoder.
In work proximal to our own, the authors of~\cite{simm2019generative} present a CVAE model based on a distance-geometric representation for the molecular conformation problem.
The generative model for molecular conformation maps atom and bond types into distances.
We map partial sets of distances and positions to full sets of distances and positions, which is equivalent to solving a distance-based formulation of the IK problem. 
Although our overall learning architecture shares similarities, we are solving a different problem.


\section{Preliminaries}
\label{sec:prelim}
Our approach casts IK as the problem of completing a partial graph of distances.
We are interested in learning how to generate full graphs conditioned on partial graphs.
In this section, we begin by introducing the general forward and inverse kinematics problems, and their solution sets, as they occur in robotic manipulation.
This is followed by a description of the distance-geometric graph representation of manipulators used in this work, where nodes correspond to points on the robot and edges correspond to known inter-point distances.
We then briefly introduce variational inference, which is the learning framework used in this work.
Finally, we describe the feature representation that we employ for learning.

\subsection{Forward and Inverse Kinematics}
\label{subsec:ik}

Most robotic manipulators are modelled as kinematic chains composed of revolute joints connected by rigid links.
The joint angles can be arranged in a vector $\Config \in \ConfigurationSpace$, where $\ConfigurationSpace \subseteq \Real^n$ is known as the manipulator \textit{configuration space}.
Analogously, coordinates $\Task$ that parameterize the task being performed constitute the \textit{task space} $\TaskSpace$.
The \textit{forward kinematics} function $FK: \ConfigurationSpace \rightarrow \TaskSpace$ maps joint angles $\Config$ to task space coordinates
\begin{equation}\label{eq:FK}
  FK(\Config) = \Task \in \TaskSpace.
\end{equation}
This relationship can be derived in closed form using known structural information (e.g., joint screws~\cite{murray2017mathematical} or Denavit-Hartenberg (DH) parameters~\cite{hartenberg1955kinematic}).
We focus on the task space $\TaskSpace := \LieGroupSE{3}$ of 6-DOF end-effector poses. 
Instead of $\Task$, we use the standard notation $\Transform$ for elements of $\LieGroupSE{3}$.

The mapping $IK: \TaskSpace \rightarrow 2^{\ConfigurationSpace}$ defines the \emph{inverse kinematics} of the robot.
In other words, $IK$ is the inverse of the forward kinematic mapping in~\cref{eq:FK}, connecting a target pose $\Transform \in \LieGroupSE{3}$ to one or more feasible configurations $\Config \in \ConfigurationSpace$.
The solution to an IK problem is generally not unique (i.e., $FK$ is not injective and therefore multiple feasible configurations exist for a single target pose $\!\Transform$).
In this paper, we consider the associated problem of determining this mapping for manipulators with $n > 6$-DOF (also known as \emph{redundant} manipulators), where each end-effector pose corresponds to a set of configurations
\begin{equation}\label{eq:IKset}
  IK(\Transform) = \left\{ \Config \in \ConfigurationSpace \, | \, FK(\Config) = \Transform \right\}
\end{equation}
that we refer to as the full set of IK solutions.
The solution set in~\cref{eq:IKset} cannot be expressed analytically for most redundant manipulators. 
Further, the infinite number of configurations cannot be recovered using local search methods.
Therefore, most of the literature refers to finding even a single feasible configuration as ``solving" inverse kinematics.
Instead of finding individual configurations that satisfy the forward kinematics equations, we approximate the full solution set $IK(\Transform)$ itself as a learned conditional distribution.
\subsection{Distance-Geometric Graph Representation of Robots}
\label{subsec:distgeo}
\begin{figure*}
  \centering
	\begin{subfigure}{0.195\textwidth}
    \centering
    \includegraphics[scale = 0.4]{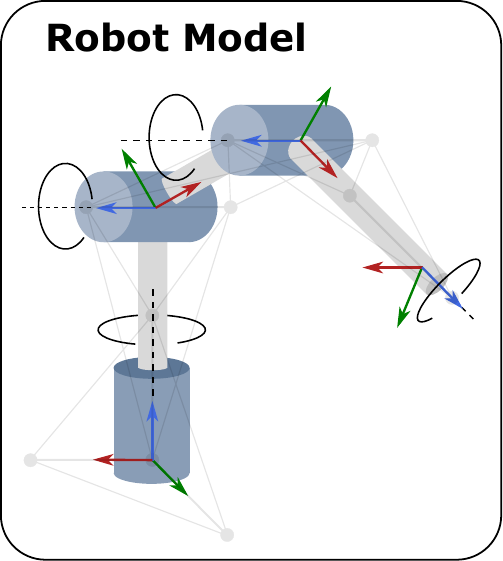}
    \caption{}\label{fig:dg_robot_model}
  \end{subfigure}
	\begin{subfigure}{0.195\textwidth}
    \centering
    \includegraphics[scale = 0.4]{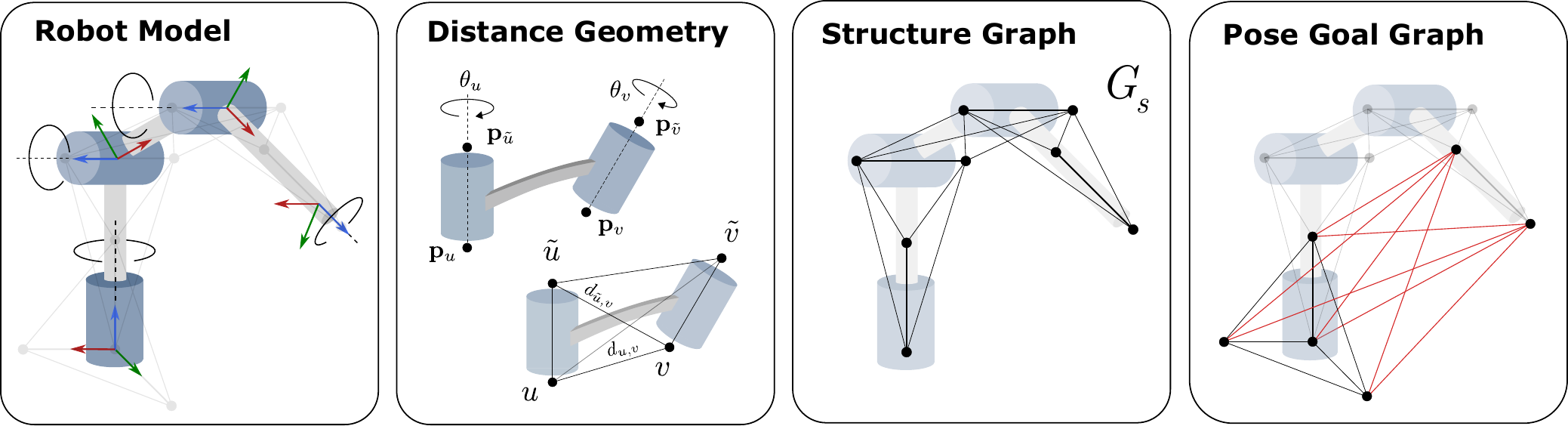}
    \caption{}\label{fig:dg_geometry}
  \end{subfigure}
	\begin{subfigure}{0.195\textwidth}
    \centering
    \includegraphics[scale = 0.4]{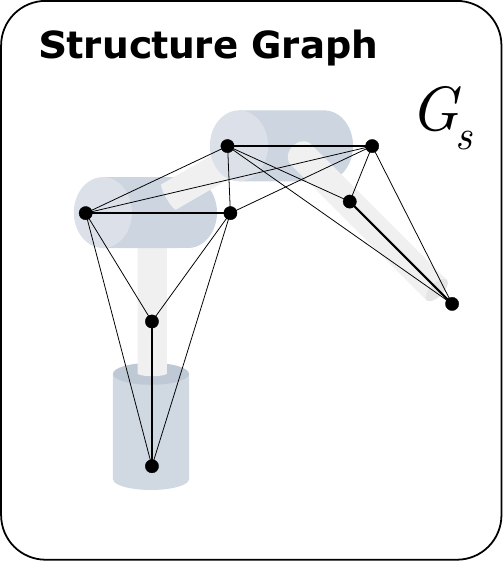}
    \caption{}\label{fig:dg_structure}
  \end{subfigure}
	\begin{subfigure}{0.195\textwidth}
    \centering
    \includegraphics[scale = 0.4]{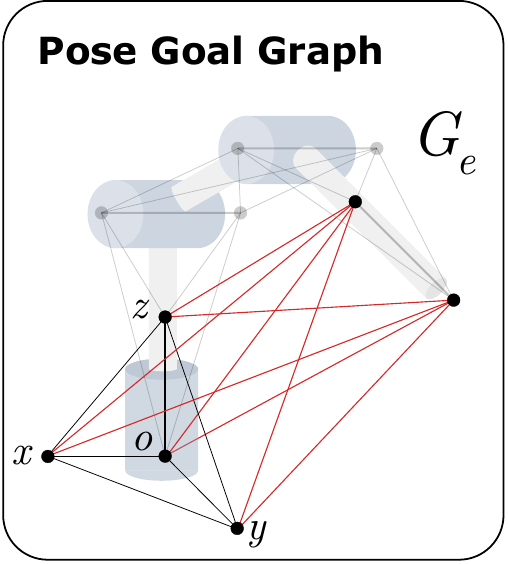}
    \caption{}\label{fig:dg_goal}
  \end{subfigure}
  	\begin{subfigure}{0.195\textwidth}
    \centering
    \includegraphics[scale = 0.4]{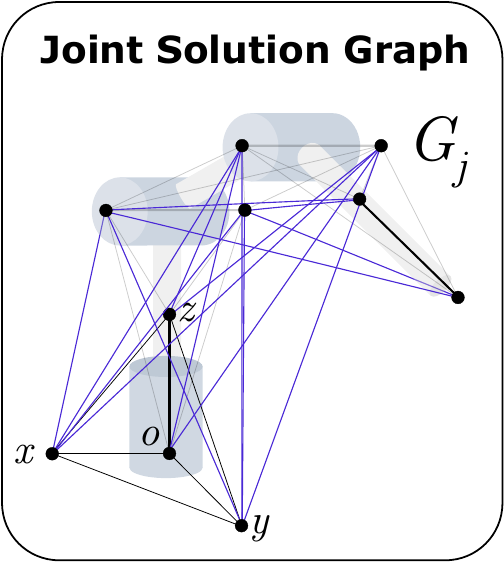}
    \caption{}\label{fig:dg_soln}
  \end{subfigure}
  \caption{The process of defining an IK problem as an incomplete or \emph{partial} graph $\widetilde{G}$ of inter-point distances and the associated IK solution as a complete graph $G$.
  (a) Conventional forward kinematics model parameterized by joint angles and joint rotation axes.
  (b) The point placement procedure for the distance-based description, first introduced in~\cite{2021_Maric_Riemannian_B}.
  Note that the six distances between points associated with pairs of consecutive joints remain constant regardless of the configuration. We annotate two out of six distances to reduce clutter.
  (c) A structure graph of the robot based on inter-point distances.
  (d) Addition of distances in red describing the robot end-effector pose using auxiliary points to define the base coordinate system, which completes the graphical IK problem description.
  All configurations of the robot reaching this end-effector pose will result in a partial graph of distances shown in (c) and (d).
  (e) The distances in blue that define a specific joint configuration.
  }
  \label{fig:dg}
\end{figure*}

We eschew the common angle-based representation of the configuration space in favour of a \emph{distance-geometric} model of robotic manipulators comprised of revolute joints~\cite{porta_branch-and-prune_2005}.
The distance-geometric approach allows us to represent a robot in configuration $\Config$ with a complete weighted graph $G=(V,E,d)$.
  The weight function $d: E \rightarrow \Real^{+}$ gives the distances between a collection of points $p$ indexed by the vertices $V$: 
  \begin{equation}
  	d_{u,v} \Defined d(\{u, v\}) = \|p(u) - p(v) \|.
  \end{equation}
  Given the complete set of distances (i.e., the range of $d$), we can use multidimensional scaling to recover unique\footnote{Up to any Euclidean transformation of $\Vector{p}$, since distances are invariant to such a transformation.} coordinates of the points $p$, which can then be used to recover the joint angles $\Config$.
  In~\cite{2021_Maric_Riemannian_B}, it was shown that the restriction $\tilde{d} \Defined d|_{\widetilde{E}}$ of the distance function $d$ to $\widetilde{E} \subset E$, represented by the incomplete or \emph{partial} graph $\widetilde{G} \Defined (V, \widetilde{E}, \tilde{d})$, may be used to recover the points $p$ corresponding to configurations that adhere to some set of geometric constraints.
  This is achieved by solving the distance geometry problem (DGP):

\begin{DGP*}[\!\cite{libertiEuclideanDistanceGeometry2014}]
\label{prob:DGP}
  Given an integer $D > 0$, a set of vertices $V$, and a simple graph $\widetilde{G} = (V,\widetilde{E}, \tilde{d})$ whose edges $\{u, v\} \in \widetilde{E}$ are assigned non-negative weights $\tilde{d}: {\{u,v\} \mapsto d_{u,v} \in \Real_{+}}$,
  find a function $ p :V \rightarrow \mathbb{R}^{D}$ such that the Euclidean distances between neighbouring vertices match their edges' weights (i.e., $\forall\,\{u, v\} \in \widetilde{E}, \, \|p(u) - p(v) \| = d_{u,v}$).
\end{DGP*}
\noindent
  Crucially, the partial graph $\widetilde{G}=(V, \widetilde{E}, \tilde{d})$ can be constructed  with $\widetilde{E} \subset E$ corresponding to distances $\tilde{d}$ determined by an end-effector pose $\Transform$ and the robot's structure.
  In this case, any set of points recovered by solving the distance geometry problem for a partial graph $\widetilde{G}$ corresponds to a particular IK solution $\Config \in IK(\Transform)$.

The generic procedure for constructing $\widetilde{G}$ is demonstrated for a simple manipulator in~\cref{fig:dg}.
We visualize the point placement step for a pair of consecutive joints in~\cref{fig:dg_geometry}. 
Two pairs of points labeled by vertices $u, \tilde{u}$ and $v, \tilde{v}$ are attached to the rotation axes of neighbouring joints at a unit distance.
The edges associated with every combination of points are then weighted by their respective distances, which are defined solely by the link geometry.
We build the \emph{structure graph} $G_{s} = (V_{s}, E_{s}, d|_{E_{s}})$ shown in~\cref{fig:dg_structure} by repeating this process for every pair of neighbouring joints.
The resulting set of vertices $V_{s}$ and distance-weighted edges $E_{s}$, shown in \cref{fig:dg_structure}, describe the rigid geometry of the robot because the edge weights are invariant to feasible motions of the robot (i.e., they remain constant in spite of changes to the configuration $\Config$).
In order to uniquely specify points with known positions (i.e., end-effectors) in terms of distances, we define the \emph{base vertices} $V_{b} = \{o, x, y, z\}$, where $o$ and $z$ are the vertices in $V_{s}$ associated with the base joint.
Setting the distances weighting the edges $E_{b}$ such that the points in $V_{b}$ form a coordinate frame with $o$ as the origin, we specify the edges $E_{p}$ weighted by distances between vertices in $V_{p} \subset V_{s}$ associated with the end-effector and the base vertices $V_{b}$.
The resulting subgraph $G_{e}= G_{b} \cup G_{p}$, where we define the union of two graphs as $G_{A} \cup G_{B} \equiv (V_{A} \cup V_{B}, E_{A} \cup E_{B}, d|_{E_{A} \cup E_{B}})$, is shown in~\cref{fig:dg_goal}. 
$G_{e}$ uniquely specifies an end-effector pose under the assumption of unconstrained rotation of the final joint, while $\widetilde{G} = G_{s} \cup G_{e}$ is the partial graph that uniquely specifies the associated IK problem.
Finally, the \emph{solution graph} $G_{j} = (V_{j}, E_{j})$ shown in~\cref{fig:dg_soln} contains the remaining distances that define a joint configuration $\Config$. 
The full graph is defined as $G = G_{s} \cup G_{e} \cup G_{j}$ and uniquely specifies a joint configuration solution to the associated IK problem~\cite{2021_Maric_Riemannian_B}.
Note that the partial graphs $\widetilde{G}$ of any two revolute manipulators with the same number of joints and goal pose differ only in the edge weights (i.e. distances) $\tilde{d}$ associated with $\widetilde{E}$, which are a function of the kinematic parameters of the two robots.
\subsection{Variational Inference}
\label{subsec:vi}
In the context of generative models, we are often interested in modelling a probability distribution $p(\Vector{x})$ from which we can draw or generate plausible samples of the multi-dimensional variable $\Vector{x}$.
A convenient and popular modelling choice is to describe the distribution $p(\Vector{x})$ as a joint distribution $p(\Vector{x}, \Vector{z})$ of the data $\Vector{x}$ and a latent variable $\Vector{z}$:
\begin{equation}\label{eq:gen_generic}
  p(\mathbf{x}) = \int p(\mathbf{x}\,|\ \mathbf{z})\, p(\mathbf{z})\, d\mathbf{z}.
\end{equation}
Directly maximizing the marginal likelihood $p(\Vector{x})$ is generally not tractable.
A common strategy in the variational inference literature \cite{Kingma2013-kw} is to instead maximize a tractable lower bound by using an approximation of the posterior $q(\Vector{z}\,|\,\Vector{x})$. 
%
We can derive this lower bound using Jensen's inequality on the log probability
%
\begin{equation}
\label{eq:elbo_derivation}
  \begin{split}
  \log p(\Vector{x}) &= \log \int p(\Vector{x}, \Vector{z})\ d\Vector{z} \\[1mm]
 &= \log \int p(\Vector{x}, \Vector{z}) \frac{q(\Vector{z} \,|\ \Vector{x})}{q(\Vector{z} \,|\ \Vector{x})}\ d\Vector{z} \\[1mm]
 &= \log \left(\Vector{E}_{q(\Vector{z}\,|\,\Vector{x})} \left[\frac{p(\Vector{x} \,|\  \Vector{z})p(\Vector{z})}{q(\Vector{z} \,|\ \Vector{x})}\right]\right)\\[1mm]
  & \geq \mathbb{E}_{q} \left[\log p(\Vector{x} \,|\  \Vector{z})p(\Vector{z}) - \log q(\Vector{z} \,|\ \Vector{x}) \right]\\[1mm]
 & \geq \mathbb{E}_{q} \left[\log p(\Vector{x} \,|\  \Vector{z}) \right] - \mathbb{E}_{q} \left[\log q(\Vector{z} \,|\ \mathbf{x}) - \log p(\Vector{z}) \right] \\[1mm]
  & \geq \mathbb{E}_{q} \left[\log p(\Vector{x} \,|\  \Vector{z}) \right] - KL( q(\Vector{z}\,|\,\Vector{x}) || p(\Vector{z})),
  \end{split}
\end{equation}
%
which is called the evidence lower bound (ELBO).
To arrive at the the common formulation of a variational auto encoder (VAE) \cite{kipf2016variational}, we parametrize the approximate posterior $q(\Vector{z} \,|\ \Vector{x})$ and the likelihood $p(\Vector{x} \,|\, \Vector{z})$ distributions with neural networks (i.e., the encoder and decoder, respectively).
Optionally, the parameters of the prior distribution $p(\Vector{z})$ may also be parametrized with a neural network.
To train all the networks, the negative ELBO is used as the loss. 
Once the network is trained, samples of $\Vector{x}$ can be generated via ancestral sampling of $p(\mathbf{z})$ and $p(\mathbf{x}\,|\ \mathbf{z})$, respectively.
A similar derivation can be performed for a conditional distribution $p(\Vector{x} \,|\ \Vector{y})$, where $\Vector{y}$ is the conditioning variable, to arrive at an ELBO defined as
\begin{equation}
\mathcal{L} = \mathbb{E}_{q(\Vector{z}\,|\,\Vector
{x}, \Vector
{y})}[\log{p(\Vector
{x}\,|\,\Vector{z}, \Vector{y})}] - KL( q(\Vector{z}\,|\,\Vector{x}, \Vector{y}) || p(\Vector{z}\,|\, \Vector{y})).
\end{equation}
If we again parametrize the (conditional) distributions with neural networks, we arrive at the CVAE formulation and loss. 

\subsection{Feature Representation for Learning}
\label{subsec:featurerep}

To train our GNN-based model, we need to choose a representation for the partial graphs $\widetilde{G}$, representing the IK problem, and complete graphs $G$, representing associated solution configurations.
For the complete graph $G$, we define the GNN node (vertex) features as a combination of point positions $\Matrix{P} = \Transpose{\left[\Vector{p}_{0}, \ldots, \Vector{p}_{N}\right]} \in \mathbb{R}^{N \times D}$, where $D \in \{2, 3\}$ is the dimension of the workspace, and general features $\Matrix{H} = \Transpose{\left[\Vector{h}_{0}, \ldots, \Vector{h}_{N}\right]} \in \mathbb{R}^{N \times L}$, where each $\Vector{h}_{i}$ is a feature vector containing extra information about the node.
For GGIK, we use a three-dimensional ($L=3$) one-hot-encoded vector for $\Vector{h}_{i}$ that indicates whether the node is part of the base coordinate system, a general joint or link, or the end-effector.
The edge features are simply the inter-point distances. We define $\epsilon_{ij} \in \mathbb{R}$ as the edge feature between node i and j.
%
For the partial graph $\widetilde{G}$, only the nodes defining the base vertices and the end-effector have known positions (i.e., the respective positions associated with nodes $V_{b}$ and $V_{p}$). The remaining unknown node positions are set to zero.
We denote $\tilde{\Vector{P}} \in \mathbb{R}^{N \times D}$ as the zero-padded node features of the partial graph.
The partial graph shares the same general features $\Matrix{H}$ as the complete graph, given that we know which part of the robot each node belongs to in advance.
%
%
The edge features of the partial graph are the known inter-point distances, with the unknown distances initialized to zero.
The distances that define the structure graph and the end-effector pose (i.e., the respective distances associated with edges $E_{s}$ and $E_{e}$) are known.
%
%
The possible choices of node and edge features offers significant flexibility.
For example, we could add an extra class or dimension to the general feature vector to indicate nodes that represent obstacles in the task space, or even more abstract features such as 3D shape information.

\section{Methodology}
\label{sec:method}

GGIK is a learning-based IK solver that is, crucially, capable of producing multiple diverse solutions while also generalizing across a family of kinematic structures. 
In this section, we cover our learning procedure (\Cref{subsec:learn}), our GNN network architecture (\Cref{subsec:architecture}), sampling procedure at test time (\Cref{subsec:sampling}) and, lastly, dataset generation and additional training and network architecture implementation details (\Cref{subsec:impl}).

\subsection{Learning to Generate Inverse Kinematics Solutions}
\label{subsec:learn}

\subsubsection{Training Procedure}

We consider the problem of modelling complete graphs corresponding to IK solutions given partial graphs that define the problem instance (i.e., the robot's geometric information and the task space goal pose).
Intuitively, we would like our network to map or ``complete'' partial graphs into full graphs.
Since multiple or infinite solutions may exist for a single end-effector pose and robot structure, a single partial graph may be associated with multiple or even infinite valid complete graphs.
We interpret the learning problem through the lens of generative modelling and treat the solution space as a multimodal distribution conditioned on a single problem instance.
By sampling from this distribution, we are able generate diverse solutions that approximately cover the space of feasible configurations.

\begin{figure*}[t]
  \centering
  \includegraphics[width=\textwidth]{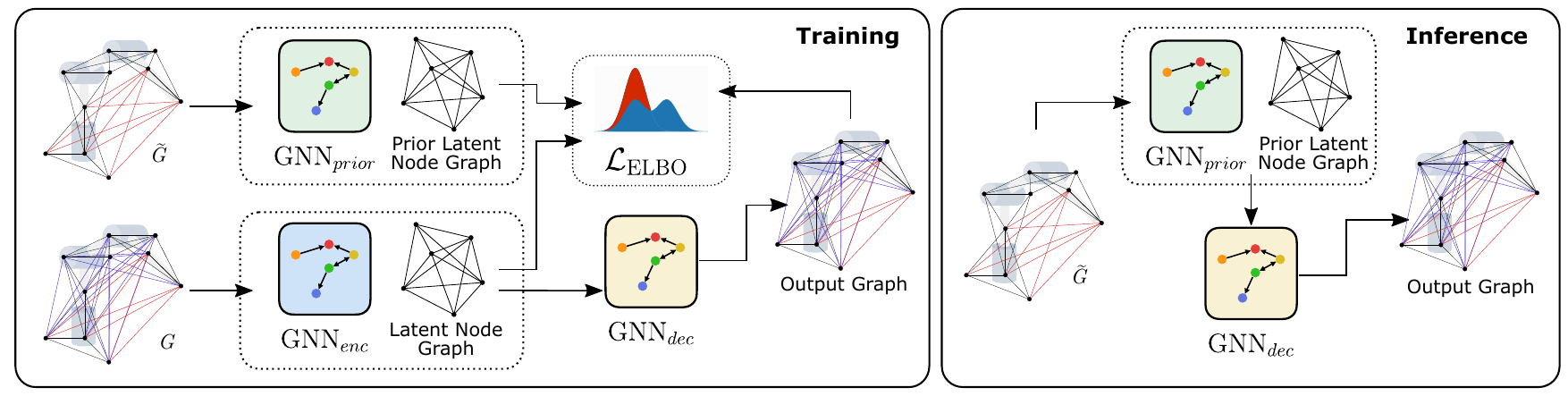}
  \caption{Our GGIK solver is based on the CVAE framework. $\text{GNN}_{enc}$ encodes a complete graph representation of a manipulator into a latent graph representation and $\text{GNN}_{dec}$ ``reconstructs" it. The prior network, $\text{GNN}_{prior}$, encodes the partial graph into a latent embedding that is near the embedding of the full graph. At test time, we decode the latent embedding of a partial graph into a complete graph to generate a solution.} \label{fig:network_architecture}
\end{figure*}

A visual overview of the training procedure is shown in~\cref{fig:network_architecture}.
At its core, GGIK is a CVAE model~\cite{sohn_cvae_2015} that parameterizes the conditional distribution $p(G\,|\,\widetilde{G})$ using GNNs.
By introducing an unobserved stochastic latent variable $\Matrix{Z}$, our generative model is defined as
\begin{equation}\label{eq:gen}
  p_{\gamma}(G\,|\,\widetilde{G}) = \int p_{\gamma}(G\,|\,\widetilde{G}, \Matrix{Z})\, p_{\gamma}(\Matrix{Z}\,|\,\widetilde{G})\, d\Matrix{Z},
\end{equation}
where $p_{\gamma}(G\,|\,\widetilde{G}, \Matrix{Z})$ is the conditional likelihood of the full graph, $p_{\gamma}(\Matrix{Z}\,|\,\widetilde{G})$ is the prior, and $\gamma$ are the learnable generative parameters. The conditional likelihood is given by
\begin{equation}\label{eq:ll}
  \begin{split}
    p_{\gamma}(G\,|\,\widetilde{G}, \Matrix{Z}) &= \prod_{i=1}^{N} p_{\gamma}(\Vector{p}_{i}\,|\,\widetilde{G}, \Vector{z}_{i}), \;\, \text{with}\\
    p_{\gamma}(\Vector{p}_{i}\,|\,\widetilde{G}, \mathbf{z}_{i}) &= \mathcal{N}(\Vector{p}_{i}\,|\,\boldsymbol{\mu}_{i}, \boldsymbol{\Sigma}_{i}),
  \end{split}
\end{equation}
where $\Matrix{Z} = \Transpose{\left[\Vector{z}_{0}, \ldots, \Vector{z}_{N}\right]} \in \mathbb{R}^{N \times F}$ are the latent embeddings of each node, and $\{ \boldsymbol{\mu}_{i}\}_{i=i}^{N}$ and $\{ \boldsymbol{\Sigma}_{i}\}_{i=i}^{N}$ are the respective predicted means and covariances of the distribution of node positions.
We parametrize the likelihood distribution with a GNN decoder, in other words, $\{ \boldsymbol{\mu}_{i}\}_{i=i}^{N}$ and $\{ \boldsymbol{\Sigma}_{i}\}_{i=i}^{N}$ are the outputs of $\text{GNN}_{dec}(\widetilde{G}, \Matrix{Z})$.
The GNN decoder propagates messages and updates the nodes at each intermediate layer and, at the final layer, outputs the predicted distribution parameters of all node positions.
In practice, for the input node features of $\text{GNN}_{dec}(\cdot)$, we concatenate the latent embeddings $\Matrix{Z}$ with the respective position features $\tilde{\Matrix{P}}$ of the original partial graph $\widetilde{G}$ and the general features $\Matrix{H}$.
%
%
The latent edge features are set as the known distances or the zero-initialized unknown distances, $\epsilon_{ij}$.
We follow the common practice of only learning the mean of the decoded Gaussian likelihood distribution and use a fixed diagonal covariance matrix $\boldsymbol{\Sigma}_{i} = \sigma_{\epsilon}\mathbf{I}$, where $\sigma_{\epsilon}$ is set to an arbitrarily small value for all $i$~\cite{doersch2016tutorial}. 
The prior distribution is given by
\begin{equation}\label{eq:prior}
  \begin{split}
    p_{\gamma}(\Matrix{Z}\,|\,\widetilde{G}) =& \prod_{i=1}^{N} p_{\gamma}(\Vector{z}_{i} \,|\,\widetilde{G}), \;\, \text{with} \\
    p_{\gamma}(\Vector{z}_{i}\,|\,\widetilde{G}) =& \sum_{k=1}^{K} \pi_{k,i}\,\mathcal{N}\left(\Vector{z}_{i}\,|\, \boldsymbol{\mu}_{k,i}, \text{diag}(\boldsymbol{\sigma}_{k,i}^{2})\right).
  \end{split}
\end{equation}

Here, we parameterize the prior as a Gaussian mixture model with $K$ components.
Each Gaussian is in turn parameterized by a mean $\boldsymbol{\mu}_{k} = \{\boldsymbol{\mu}_{k,i}\}_{i=1}^{N}$, diagonal covariance $\boldsymbol{\sigma}_{k} = \{\boldsymbol{\sigma}_{k,i}\}_{i=1}^{N}$, and a mixing coefficient $\boldsymbol{\pi}_{k} = \{\pi_{k, i}\}_{i=1}^{N}$, where $\sum_{k=1}^{K} \pi_{k, i} = 1,\,\, \forall\, i=1, ..., N$.
We chose a mixture model to have an expressive prior capable of capturing the latent distribution of multiple solutions.
We parameterize the prior distribution with a multi-headed GNN encoder $\text{GNN}_{prior}(\widetilde{G})$ that outputs parameters $\{\boldsymbol{\mu}_{k}, \boldsymbol{\sigma}_{k}, \boldsymbol{\pi}_{k}\}_{k=1}^{K}$.

For GGIK, the goal of learning is to maximize the marginal likelihood or evidence of the full graphs as shown in \cref{eq:gen}.
As detailed in \Cref{subsec:vi}, we do this by maximizing the ELBO
  \begin{align} \label{eq:elbo}
	& \mathcal{L} = \mathbb{E}_{q_{\phi}(\mathbf{z}\,|\,G)}[\log{p_{\gamma}(G\,|\,\widetilde{G}, \Matrix{Z})}] - \nonumber \\ 
	& \hspace{35pt} \beta KL( q_{\phi}(\Matrix{Z}\,|\,G) || p_{\gamma}(\Matrix{Z}\,|\, \widetilde{G})),
  \end{align}
where $KL(\cdot||\cdot)$ is the Kullback-Leibler (KL) divergence, $\beta$ is a weighting hyperparameter, and $q_{\phi}(\Matrix{Z}\,|\,G)$ is the inference model with learnable parameters $\phi$, defined as
\begin{equation}\label{eq:inf}
  \begin{split}
	q_{\phi}(\Matrix{Z}\,|\,G) = \prod_{i=1}^{N} q_{\phi}(\Vector{z}_{i}\,|\,G), \quad \text{with} \\ 
	q_{\phi}(\Vector{z}_{i}\,|\,G) = \mathcal{N}(\Vector{z}_{i}\,|\, \boldsymbol{\mu}_{i}, \text{diag}(\boldsymbol{\sigma}_{i}^{2})).
  \end{split}
\end{equation}
As with the prior distribution, we parameterize the inference distribution with a multi-headed GNN encoder, $\text{GNN}_{enc}(G)$, that outputs parameters $\{ \boldsymbol{\mu}_{i}\}_{i=1}^{N}$ and $\{ \boldsymbol{\sigma}_{i}\}_{i=1}^{N}$.
The inference model is an approximation of the intractable true posterior $p(\Matrix{Z}\,|\,G)$.
We note that the resulting ELBO objective in \cref{eq:elbo} is based on an expectation with respect to the inference distribution $q_{\phi}(\Matrix{Z}\,|\,G)$, which itself is based on the parameters $\phi$.
Since we restrict $q_{\phi}(\Matrix{Z}\,|\,G)$ to be a Gaussian variational approximation, we can use stochastic gradient descent (i.e., Monte Carlo gradient estimates) via the reparameterization trick~\cite{Kingma2013-kw} to optimize the lower bound with respect to parameters $\gamma$ and $\phi$.

\subsubsection{Inference Procedure}
At test time, given a goal pose and the manipulator's geometric information encapsulated in a partial graph $\widetilde{G}$, the prior network, $\text{GNN}_{prior}$, encodes the partial graph as a latent distribution $p_{\gamma}(\Matrix{Z}\,|\,\widetilde{G})$.
During training, the distribution $p_{\gamma}(\Matrix{Z}\,|\,\widetilde{G})$ is optimized to be simultaneously near multiple encodings of valid solutions by the KL divergence term in~\cref{eq:elbo}.
We can sample this multimodal distribution $\Matrix{Z} \sim p_{\gamma}(\Matrix{Z}\,|\, \widetilde{G})$ as many times as needed, and subsequently decode all of the samples with the decoder network $\text{GNN}_{dec}$ to generate IK solutions represented as complete graphs.
This procedure can be done quickly and in parallel on the GPU.
We provide a more detailed explanation of the sampling procedure in~\cref{subsec:sampling}. 
%

\subsection{Equivariant Network Architecture}
\label{subsec:architecture}
In this section, we discuss the choice of architecture for networks $\text{GNN}_{dec}$, $\text{GNN}_{enc}$, and $\text{GNN}_{prior}$.
%
%
During inference, our model maps zero-padded partial point sets $\tilde{\Matrix{P}}$ to full point sets $\Matrix{P}$, $f\,:\,\mathbb{R}^{N \times D} \rightarrow  \mathbb{R}^{N \times D}$, where $f$ is a combination of networks $\text{GNN}_{prior}$ and $\text{GNN}_{dec}$ applied sequentially.
The point positions of the nodes in the distance geometry problem contain underlying geometric relationships and symmetries that we would like to preserve with our choice of architecture.
Most importantly, the point sets are \emph{equivariant} to the Euclidean group $\LieGroupE{n}$ of rotations, translations, and reflections.
Let $S: \mathbb{R}^{N \times D} \rightarrow \mathbb{R}^{N \times D}$ be a transformation consisting of some combination of rotations, translations and reflections on the initial zero-padded partial point set $\tilde{\Matrix{P}}$.
Then, there exists an equivalent transformation $T: \mathbb{R}^{N \times D} \rightarrow \mathbb{R}^{N \times D}$ on the complete point set $\Matrix{P}$ such that
\begin{equation}\label{eq:equivariance}
f(S(\tilde{\Matrix{P}})) = T(f(\tilde{\Matrix{P}})).
\end{equation}
%
As a specific example, if we translate and rotate our initial partial graph, which defines the IK problem instance, then the complete graph, which defines the solution, will be equivalently translated and rotated.
To leverage this geometric prior or structure in the data, we use $\LieGroupE{n}$-equivariant graph neural networks (EGNNs)~\cite{satorras2021n} for $\text{GNN}_{dec}$, $\text{GNN}_{enc}$, and $\text{GNN}_{prior}$.
The EGNN layer splits up the node features into an equivariant coordinate or position-based part and a non-equivariant part.
We treat the positions $\Matrix{P}$ and $\tilde{\Matrix{P}}$ as the equivariant portion and the general features $\Matrix{H}$ as non-equivariant.
As an example, for a single EGNN layer $l$ and a single node $i$, the update equations for the equivariant node features are
\begin{equation}\label{eq:egnn_layer_p1}
  \begin{split}
  	\mathbf{p}_{i}^{l+1} &= \mathbf{p}_{i}^{l} + C \sum_{j \neq i} (\mathbf{p}_{i}^{l} - \mathbf{p}_{j}^{l}) \phi_{x}(\mathbf{m}_{ij}) \\
	\mathbf{m}_{ij} &= \phi_{e}(\mathbf{h}_{i}^{l}, \mathbf{h}_{j}^{l}, \| \mathbf{p}_{i}^{l} - \mathbf{p}_{j}^{l}\|^{2}, \epsilon_{ij}),
  \end{split}
\end{equation}
where we define $\mathbf{m}_{ij} \in \mathbb{R}^{d}$ as a message of dimension $d$ from node $j$ to $i$. 
The position of each node is updated by the weighed sum of relative differences between the node itself and all other nodes.
The function $\phi_{x}: \mathbb{R}^{d} \rightarrow \mathbb{R}$ provides the weights for each term in the sum.
The function $\phi_{e}$ represents a general edge operation.
$C$ is a normalizing factor based on the number of elements. 
Similarly, the update equations for the non-equivariant node features are
\begin{equation}\label{eq:egnn_layer_p2}
  \begin{split}
	\mathbf{h}_{i}^{l+1} &= \phi_{h}(\mathbf{h}_{i}^{l}, \mathbf{m}_{i}) \\
	\mathbf{m}_{i} &= \sum_{j \neq i} \mathbf{m}_{ij},
  \end{split}
\end{equation}
where $\phi_{h}$ represents a node update operation and $\mathbf{m}_{i} \in \mathbb{R}^{d}$ is the sum of all messages received by node $i$.
We model $\phi_{x}$, $\phi_{h}$, and $\phi_{e}$ with multilayer perceptrons (MLPs).
%
For all learned distributions, we map the concatenated equivariant and non-equivariant features of each node from the last layer to the respective distribution parameters.
For more details about the model and a proof of the equivariance property, we refer readers to~\cite{satorras2021n}.
We present ablation studies on the use of the EGNN network architecture in Section \ref{sec:exp}, demonstrating its importance to our approach.
\subsection{Sampling Inverse Kinematics Solutions}
\label{subsec:sampling}

\begin{algorithm}[b] 
	\SetAlgoLined
	\Parameters{$\widetilde{G}, \Matrix{T}_{goal}, N, M$}
	\KwResult{Solution configurations with the lowest pose error $\Config^{*} \in \Real^{M \times n_{joints}}$.}
  \vspace{1mm}
  $\Matrix{Z}_{N} \sim p_{\gamma}(\Matrix{Z}\,\vert\,\widetilde{G})$\hspace*{-0.7mm} \Comment*[f]{Sample $N$ latents $\Matrix{Z}$ from $\text{GNN}_{prior}$.} \\
  \vspace{1mm}
  $\Matrix{P}_{N} \sim p_{\gamma}(\Matrix{P}\,\vert\,\widetilde{G}, \Matrix{Z}_{N} )$ \Comment*[f]{Get $N$ solutions via $\text{GNN}_{dec}$.} \\
  \vspace{1mm}
  $\Config_{N} \leftarrow \text{fromPoints}(\Matrix{P}_{N})$ \Comment*[f]{Recover $N$ configurations.} \\
  \vspace{1mm}
  $\Config^{*} \leftarrow \text{selectSol}(\Matrix{T}_{goal}, \Config_{N}, M)$ \Comment*[f]{Choose best $M$.}\\
	\caption{GGIK}
	\label{alg:GGIK}
\end{algorithm}

We summarize the full sampling procedure in~\cref{alg:GGIK}.
We first sample $N$ sets of latent graphs $\Matrix{Z}_{N}$ from a Gaussian distribution $p_{\gamma}(\Matrix{Z}\,\vert\,\widetilde{G})$ parameterized by outputs from the prior encoder $\text{GNN}_{prior}$.
The samples $\Matrix{Z}_{N}$ are then passed to the decoder network $\text{GNN}_{dec}$, producing $N$ point sets $\Matrix{P}_{N}$ sampled from the distribution $p_{\gamma}(\Matrix{P}\,\vert\,\widetilde{G}, \Matrix{Z})$.
Next, $N$ joint configuration (i.e., joint angle) estimates $\Config_{N}$ are recovered from the point sets $\Matrix{P}_{N}$ using a simple geometric procedure described in~\cite{2021_Maric_Riemannian_B}.
Finally, we select the $M$ best configurations $\Config^{*}$ according to some criteria.
We choose to filter for the $M$ configurations with the lowest error with respect to the goal pose $\Matrix{T}_{goal}$, where we define the error as
\begin{equation*}
e = \Norm{\MatLog{\Matrix{T}^{-1}\Matrix{T}_{goal}}}_{2}.
\end{equation*}
Note that the configuration recovery and filtering steps may be performed in parallel for all $N$ configurations, in which case they require approximately 5 ms of computation time on a laptop CPU for $N =$ 128 and $M =$ 32.
For more details on the total computation time required, see \cref{subsec:expAppendix1}.
\subsection{Learning Implementation Details}
\label{subsec:impl}

In this section, we first cover how we generate our training dataset, followed by additional training and network architecture details.

\subsubsection{Training Data}
\label{subsec:datagen}
Acquiring training data is convenient---any valid manipulator configuration can be used for training.
In our experiments, we use training data from two different datasets.
The first dataset comprises of IK problems for existing commercial manipulators with typical kinematic structures.
The second dataset contains IK problems for manipulators with randomly generated kinematic structures.
In both datasets, IK ``instances" were generated by and uniformly sampling a set of joint angles and using forward kinematics to compute the associated end-effector pose.
The end-effector pose is then used as the goal for IK, which ensures that all problems are feasible (i.e., have at least one solution).

\begin{figure*}[]
  \centering
  \begin{subfigure}{0.195\textwidth}
    \includegraphics[width=0.8\linewidth]{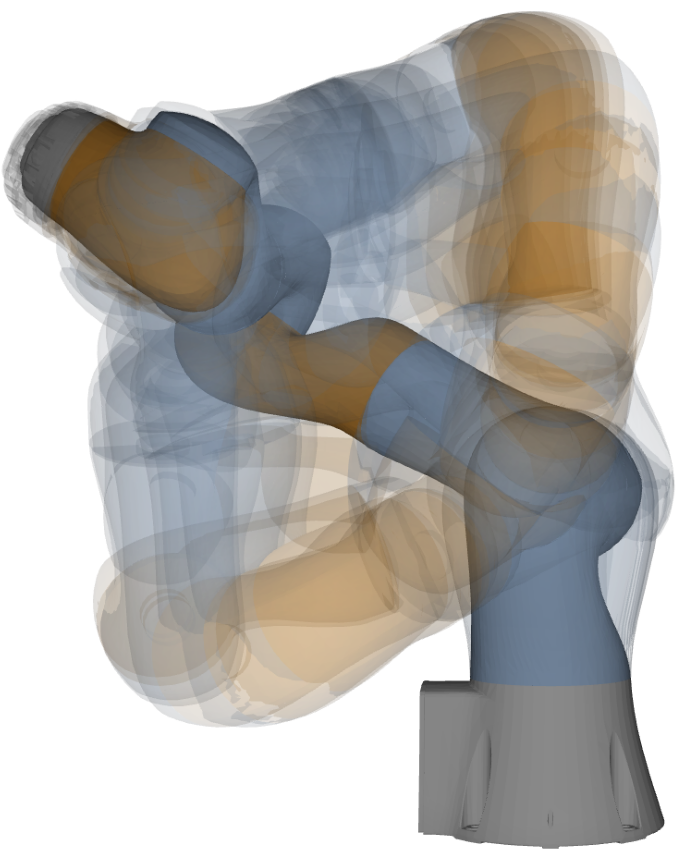}
  \end{subfigure}
  \begin{subfigure}{0.195\textwidth}
    \includegraphics[width=0.8\linewidth]{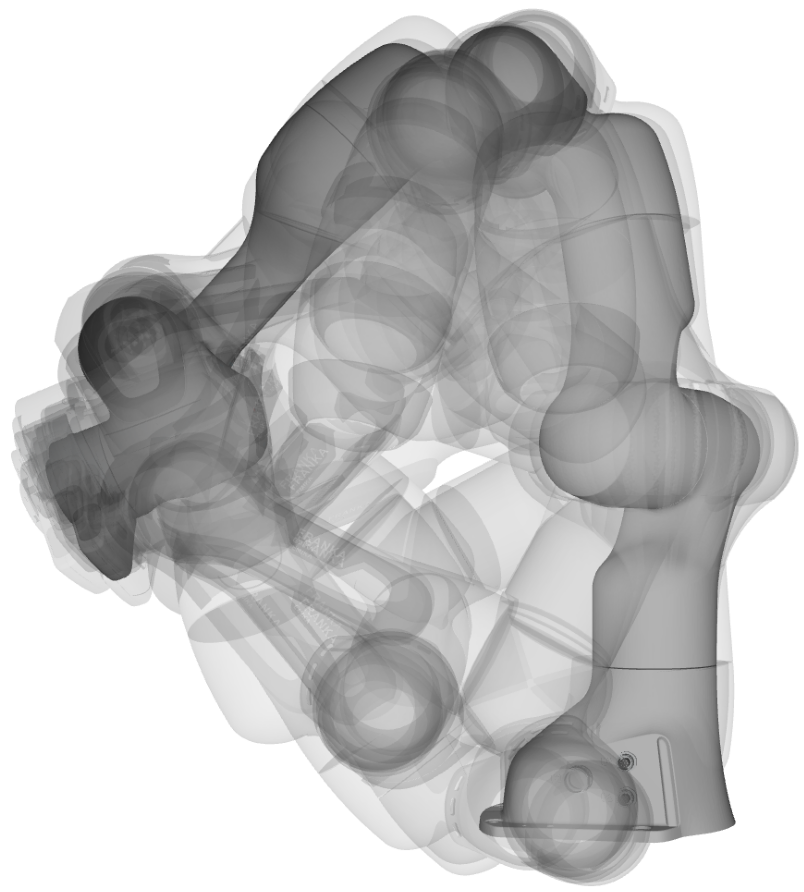}
  \end{subfigure}
  \begin{subfigure}{0.195\textwidth}
    \includegraphics[width=0.8\linewidth]{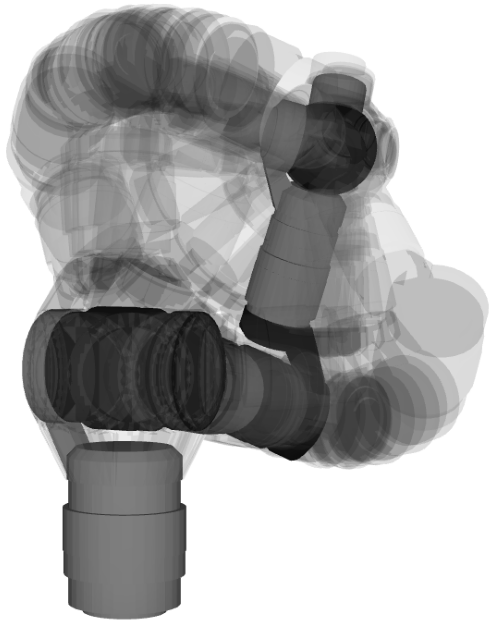}
  \end{subfigure}
  \begin{subfigure}{0.195\textwidth}
    \includegraphics[width=0.8\linewidth]{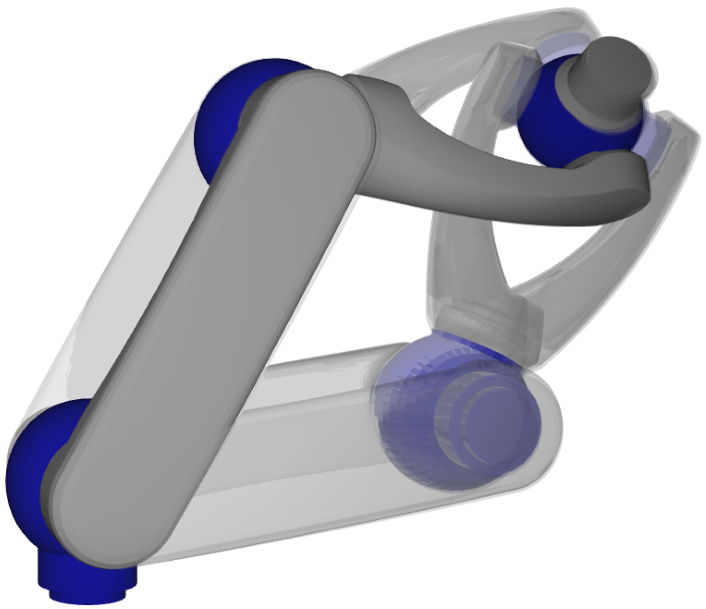}
  \end{subfigure}
  \begin{subfigure}{0.195\textwidth}
    \includegraphics[width=0.8\linewidth]{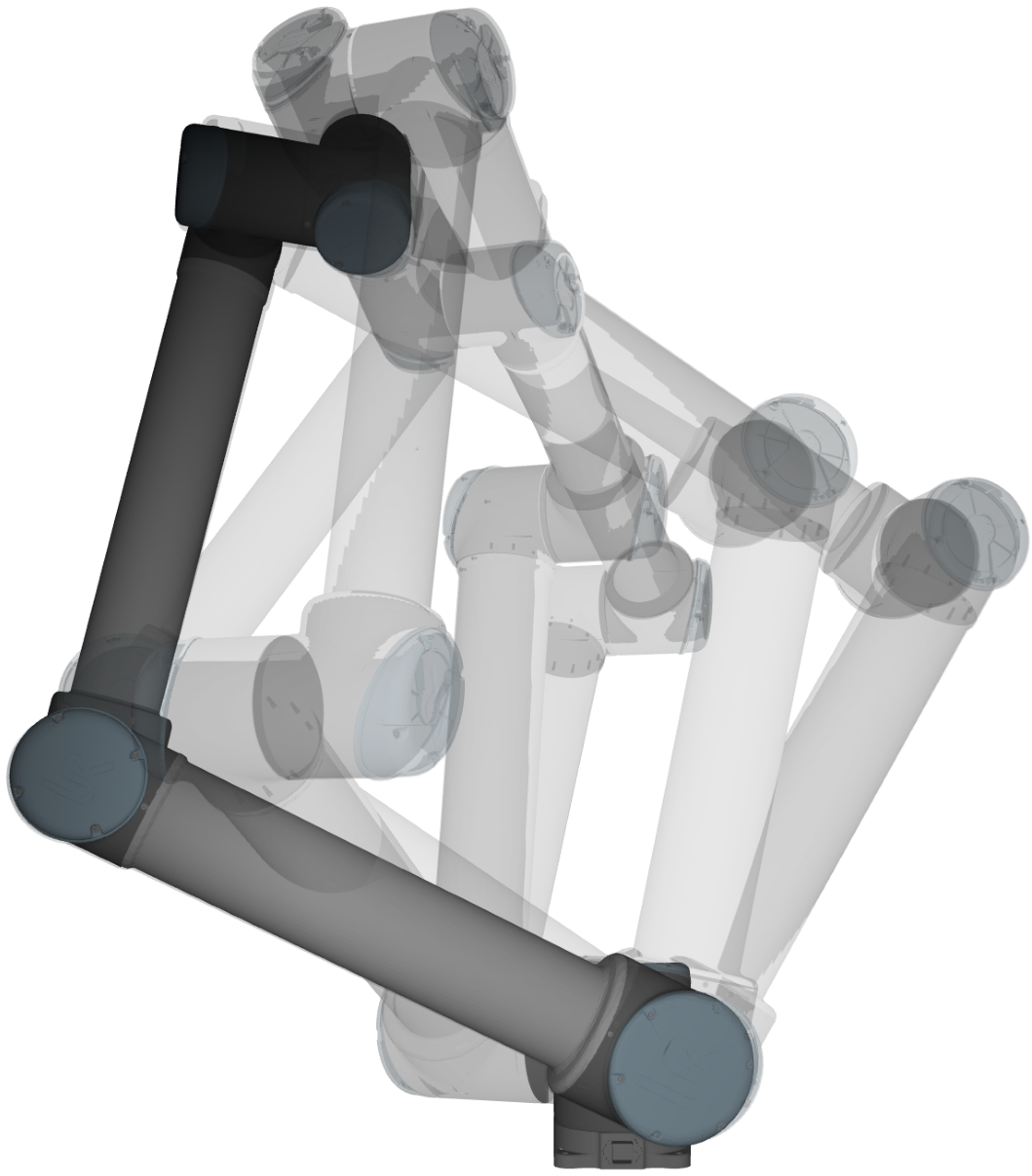}
  \end{subfigure}%
  \vspace{2mm}
  \caption{Sampled conditional distributions from GGIK for various robotic manipulators produced by a single model. From left to right: KUKA IIWA, Franka Emika Panda, Schunk LWA4D, Schunk LWA4P, and Universal Robots UR10. Note that the end-effector poses are nearly identical in all cases, highlighting kinematic redundancy. Furthermore, the discrete solution sets of the two 6-DOF robots are captured by our model also.}
  \label{fig:exp2}
\end{figure*}

\paragraph{Commercial Manipulator Dataset} This dataset contains IK problems that are uniformly distributed over five different robots with varying kinematic properties.
Specifically, we chose the Universal Robots UR10, Schunk LWA4P, Schunk LWA4D, KUKA IIWA and the Franka Emika Panda robots.
The UR10 and LWA4P are 6-DOF robots with up to 8 discrete IK solutions for a given goal pose.
On the other hand, the IIWA, LWA4D and the Panda robots all share a redundant kinematic structure with 7-DOF and, by extension, an infinite solution set to any feasible IK problem.
The kinematic models for these robots are computed by parsing their respective URDF files, which allows us to render solutions generated by our model, as shown in~\cref{fig:exp2}.

\paragraph{Randomized Manipulator Dataset} This dataset contains a large number of randomly generated pairs of robots and associated IK problems.
A unique robot was created for each problem.
A key consideration for any randomization scheme used to produce such a dataset is that most real manipulators exist within a narrow class of kinematic structures.
For example, the rotation axes of consecutive joints for many manipulators are either parallel or perpendicular.
Furthermore, manipulators generally do not feature more than two sequential joints with parallel rotation axes, and these joints are generally displaced along a direction orthogonal to their axis of rotation.
Therefore, we generate samples by randomizing DH parameters following the original convention of \cite{hartenberg1955kinematic} using the procedure in~\cref{alg:rDH}.
%
  %
  In the absence of an informative prior, we sample from a uniform distribution that can be discrete (e.g., $\mathcal{U}\{a,b,c,\cdots\}$) or continuous within a range (e.g., $\mathcal{U}(a,b)$).
  We select the parameters of this distribution such that the resulting structures include the UR10, KUKA, LWA4D and LWA4P robot, although it is improbable that the exact parameters of these robots will be sampled.
%
The resulting problems are constrained to robots with structures similar, but never identical to, the commercial robots used in the first dataset.
\begin{algorithm}[t]
	\SetAlgoLined
	\Parameters{$d_{min}, d_{max}, a_{min}, a_{max}$}
	\KwResult{Randomized robot}
  $\alpha_0 \sim \mathcal{U}\{ -\pi /2, \pi /2 \} $ \Comment*[f]{First set of DH parameters.} \\
  $a_0, \theta_{0} \leftarrow 0$ \\
  $d_0 \sim \mathcal{U}\left(0, d_{max}\right)$ \\
	\For {$k = 1,\,\dots\, n-2$}{
    $\theta_{k}, a_{k}, d_{k} \leftarrow 0$ \\
        \Comment{Limit sequential parallel rotation axes.}
        \uIf{$\alpha_{k-1} = 0$ and $\alpha_{k-2} = 0$}{
        $ \alpha_k \sim \mathcal{U}\{ -\pi /2, \pi /2 \} $ \\
            }
        \Else{
        $\alpha_k \sim \mathcal{U}\{ 0, -\pi /2, \pi /2 \} $ \\
        }
        \Comment{Ensure rotation axes are coplanar.}
        \uIf{$\alpha_{k} \neq 0$}{
        $d_k \sim \mathcal{U}\left(0, d_{max}\right)$ \\ 
        }
        \Else{
        $a_k \sim \mathcal{U}\left(a_{min}, a_{max}\right)$ \\
        }
	}
  $a_{n-1}, \theta_{n-1}, \alpha_{n-1} \leftarrow 0 $ \Comment*[f]{End-effector frame.} \\
  $d_{n-1} \sim \mathcal{U}\left(0, d_{max}\right)$ \\
	\caption{Randomized DH parameters}
	\label{alg:rDH}
\end{algorithm}

\subsubsection{Additional Training and Network Architecture Details}
We provide some brief and practical advice for training GGIK based on our experimental observations.
Picking a network architecture that captures the underlying geometry and equivariance of the problem improved the performance of the models by a large margin, as shown by \cref{table:exp4}.
%
%
We used 16 components for the Gaussian mixture model parameters from $\text{GNN}_{prior}$.
The MLPs used for message passing in the EGNN: $\phi_{e}$, $\phi_{h}$, and $\phi_{x}$ all had two hidden layers of dimension 128.
In order to increase the accuracy of our models, we found it particularly effective to lower the learning rate when the training loss stagnated.
We empirically observed that increasing the number of layers used in the GNNs up to five improved the overall performance.
There was very little performance improvement beyond five layers.
We found that using a mixture model for the distribution of the prior encoder lead to more accurate results.
We hypothesize that a more expressive prior distribution can better fit multiple encoded IK solutions when conditioned on a single goal pose.
Finally, we found that lowering the weight of the reconstruction loss or likelihood contribution from nodes that belonged to the \emph{structure graph} (see~\cref{subsec:distgeo}) led to overall lower losses. 
We hypothesize that by lowering the weight of the reconstruction loss for the structure graph, we encourage the optimization procedure to focus on reconstructing non-trivial parts of the solution, which then avoids local minima that overemphasize learning the structure graph $G_s$.
We used SiLU non-linearities~\cite{elfwing2018sigmoid} with layer normalization~\cite{ba2016layer} for all networks. 
We used the Adam optimizer with weight decay~\cite{LoshchilovH19} and a learning rate of 3e-4.
The smaller models were each trained for a total of approximately 16 hours on 512,000 IK problems and the larger models were trained for approximately 90 hours on 2,560,000 IK problems.
All models were trained for 360 epochs.
For more implementation details, we invite interested readers to refer to our open-source code repository.
%

\section{Experiments}
\label{sec:exp}
In this section, we present results from a series of seven experiments, where we studied the behaviour and properties of GGIK. We evaluated (i) GGIK's capability to learn accurate solutions and generalize within a class of manipulator structures (\Cref{subsec:exp1}), (ii) GGIK's ability to generalize to unseen manipulators (\Cref{subsec:generalization}), (iii) GGIK's ability to produce multiple solutions (\Cref{subsec:multimod}), (iv) the scaling of the computation times of GGIK (\Cref{subsec:expAppendix1}), (v) whether GGIK can effectively initialize local numerical IK solvers (\Cref{subsec:exp2}), (vi) the importance of our choice of learning architecture (\Cref{subsec:exp4}), and (vii) whether or not GGIK can learn constrained distributions that account for joint limits (\Cref{subsec:expjointlimits}).
%
%
All experiments were performed on a computer with an AMD Ryzen Threadripper 2920X 12-Core Processor CPU and an NVIDIA Quadro RTX 8000 GPU.

\subsection{Accuracy}
\label{subsec:exp1}

\begin{table*}
  \centering
  \input{tables/experiment_2_individual.tex}
  \caption{Performance of GGIK instances on 2,000 randomly generated IK problems per individual robotic manipulator types. A separate GGIK model was trained for each of the five robots. For GGIK, taking 32 samples from the learned distribution, the error statistics are presented as the mean and mean minimum and maximum error per problem, as well as the two quartiles of the distribution. We include baseline results from distal teaching (DT) \cite{vonoehsen_comparison_2020}, IKFlow \cite{ames2021ikflow} and IKNet \cite{2022_Bensadoun_Neural}. Dashed results were not provided by previous work. For the DT baseline, we calculate the statistics directly over all 2,000 problems.
  }
\label{table:exp2_individual}
\end{table*}

\begin{table*}
  \centering

\input{tables/experiment_3.tex}
  \caption{Performance of GGIK on 10,000 randomly generated IK problems (2,000 per robot) for a single model trained on five different robotic manipulators. Taking 32 samples from the learned distribution, the error statistics are presented as the mean and mean minimum and maximum error per problem, as well as the two quartiles of the distribution. Note that all solutions were produced by a \emph{single} model.}
\label{table:exp2}
\end{table*}

We evaluated the accuracy of GGIK for a variety of existing commercial manipulators featuring different structures and numbers of joints: the Kuka IIWA, Schunk LWA4D, Schunk LWA4P, Universal Robots UR10, and Franka Emika Panda.
We also assessed the performance of GGIK for a hypothetical revolute robot with 12-DOF, to demonstrate the feasibility of scaling to more complex manipulators.

\subsubsection{Individual Robots} We trained a bespoke model for each of the robots on 512,000 data points (IK problems), which we generated using the procedure described in~\cref{subsec:datagen}.
We evaluated these models on 2,000 randomly-generated IK problem instances per robot and report the error between the computed and goal end-effector poses. 
The orientation error is the magnitude of the angular difference in orientation, measured using the axis-angle representation.
The results in~\cref{table:exp2_individual} show that GGIK generates solutions with a mean position error of under 6 mm and a mean orientation error of under 0.4 degrees.
Moreover, the first ($Q_{1}$) and third ($Q_{3}$) quartiles of the error distributions, as well as the maximum error values, show that the majority of sampled configurations produced end-effector poses with errors that do not substantially deviate from the mean.
%
%
%
Overall, these results indicate that, when trained on a particular manipulator, GGIK has the ability to produce solutions with an accuracy sufficient for common manipulation tasks such as grasping, pushing, and obstacle avoidance.

As a benchmark, we trained a comparably sized network on 512,000 UR10 poses using the distal teaching (DT) method of \cite{vonoehsen_comparison_2020}.
The DT network was only able to achieve a mean position error of 43.5 mm and a mean rotation error of 15.3 degrees.
Importantly, this network maps an end-effector pose to a single fixed solution and does not model the entire solution set---multiple calls to the network will always produce the same solution.
We also compare the performance of GGIK with results reported from the IKFlow~\cite{ames2021ikflow} and IKNet~\cite{2022_Bensadoun_Neural} models on the Franka Emika Panda manipulator. 
GGIK achieved better overall accuracy than IKFlow with approximately 2 mm lower translational error and 2 degrees lower rotational error on average.
It is also important to note that the IKFlow model was trained on 2.5 million problems per robot, roughly five times more than GGIK (see \cref{table:exp2_individual}).
GGIK achieves significantly better accuracy than IKNet, with approximately 25 mm less translational error on average.
The authors of IKNet did not report on rotational errors or the size of their training set.

\subsubsection{Multiple Robots}
\label{subsec:mrob}
Next, we trained a single instance of GGIK on a total of 2,560,000 IK problems uniformly distributed over all five manipulators, that is, a single model for multiple robots.
The model was evaluated on 10,000 randomly generated feasible IK problems (2,000 per robot).
The error rates in~\cref{table:exp2} suggest that the proposed approach can be used to generate solutions in a variety of practical applications.
%
%
The results shown in~\cref{fig:boxplots_planar} indicate that by simply increasing the number of samples, the gap in solution accuracy between the various manipulators is reduced.

\begin{figure}
	\centering
	\begin{subfigure}[t]{\columnwidth}
  \centering
    \begin{adjustbox}{width=0.9\linewidth}
      \begin{tikzpicture}[font=\sffamily\small]
        \pgfplotsset{
          width=\linewidth,
          ylabel style={yshift=-5pt},
          xlabel style={yshift=5pt},
          yticklabel style={font=\small},
          legend style={font=\small},
		  compat=1.11,
          /pgfplots/ybar legend/.style={
          /pgfplots/legend image code/.code={%
          \draw[##1,/tikz/.cd,yshift=-0.25em]
          (0cm,0cm) rectangle (3pt,0.8em);},
         },
        }
	    \tikzstyle{every node}=[font=\small]
        \input{figures/tikz/box_comp_pos.tex}
      \end{tikzpicture}
    \end{adjustbox}
		\caption{Position error}\label{fig:boxplot_all_pos}
    \vspace{2mm}
	\end{subfigure}
	\begin{subfigure}[t]{\columnwidth}
		\centering
		\begin{adjustbox}{width=0.9\linewidth}
      \begin{tikzpicture}
        \pgfplotsset{
          width=\linewidth,
          ylabel style={yshift=-5pt},
          xlabel style={yshift=5pt},
          yticklabel style={font=\small},
          legend style={font=\small},
        }
        \input{figures/tikz/box_comp_rot.tex}
      \end{tikzpicture}
		\end{adjustbox}
		\caption{Rotation error}\label{fig:boxplot_all_rot}
	\end{subfigure}
	\caption{Box-and-whiskers plots comparing the accuracy for identical models trained on datasets containing multiple robots distributed over 512,000 and 2,560,000 datapoints, respectively.}
	\label{fig:boxplots_planar}
\end{figure}

\subsection{Generalization}
\label{subsec:generalization}
\begin{table*}
  \centering

\input{tables/experiment_x.tex}
  \caption{Performance of GGIK on manipulators not seen during training. We train on randomly generated manipulators of DOF 6 and 7, and test on actual commercial robots, which were not seen in the dataset. Taking 50 samples from the learned distribution, the error statistics are presented as the mean and mean minimum and maximum error per problem, as well as the two quartiles of the distribution.}
  \label{table:ood}
\end{table*}

In order to determine the capacity of GGIK to generalize to unseen manipulator structures, we evaluated the accuracy of a trained model on robots outside of the training set.
To this end, we generated a dataset of inverse kinematics problems for manipulators with randomized kinematic structures, following the procedure outlined in \cref{subsec:datagen}.
The training set consists of 4,096,000 inverse kinematics problems, each for a unique 6- or 7-DOF robot generated using randomized DH parameters (i.e., no two robots in the dataset are identical).
We randomly selected batches of 128 robots during training, independent of their DOF and structure.
The robots and procedure used in evaluation are identical to the experiment presented in~\cref{subsec:exp1}.
Each line in~\cref{table:ood} displays the results from 2,000 IK problems solved by the trained model for a particular robot.

Our learned model was able to generalize reasonably well across all robots, reaching a mean position error of approximately 2 to 4 cm and a mean orientation error of 2 degrees.
In contrast to the model trained on specific robots, discussed in~\cref{subsec:exp1}, this model exchanges accuracy for the capability to generate approximate solutions for a much larger variety of robots.
While the accuracy of this model is not sufficient alone for some practical use cases, the results in \Cref{subsec:exp2} suggest that the approximate solutions it provides can be refined by a few iterations of a local optimization procedure.

\subsection{Multimodality}
\label{subsec:multimod}

\begin{figure*}[t]
  \centering
  \includegraphics[width=0.95\textwidth]{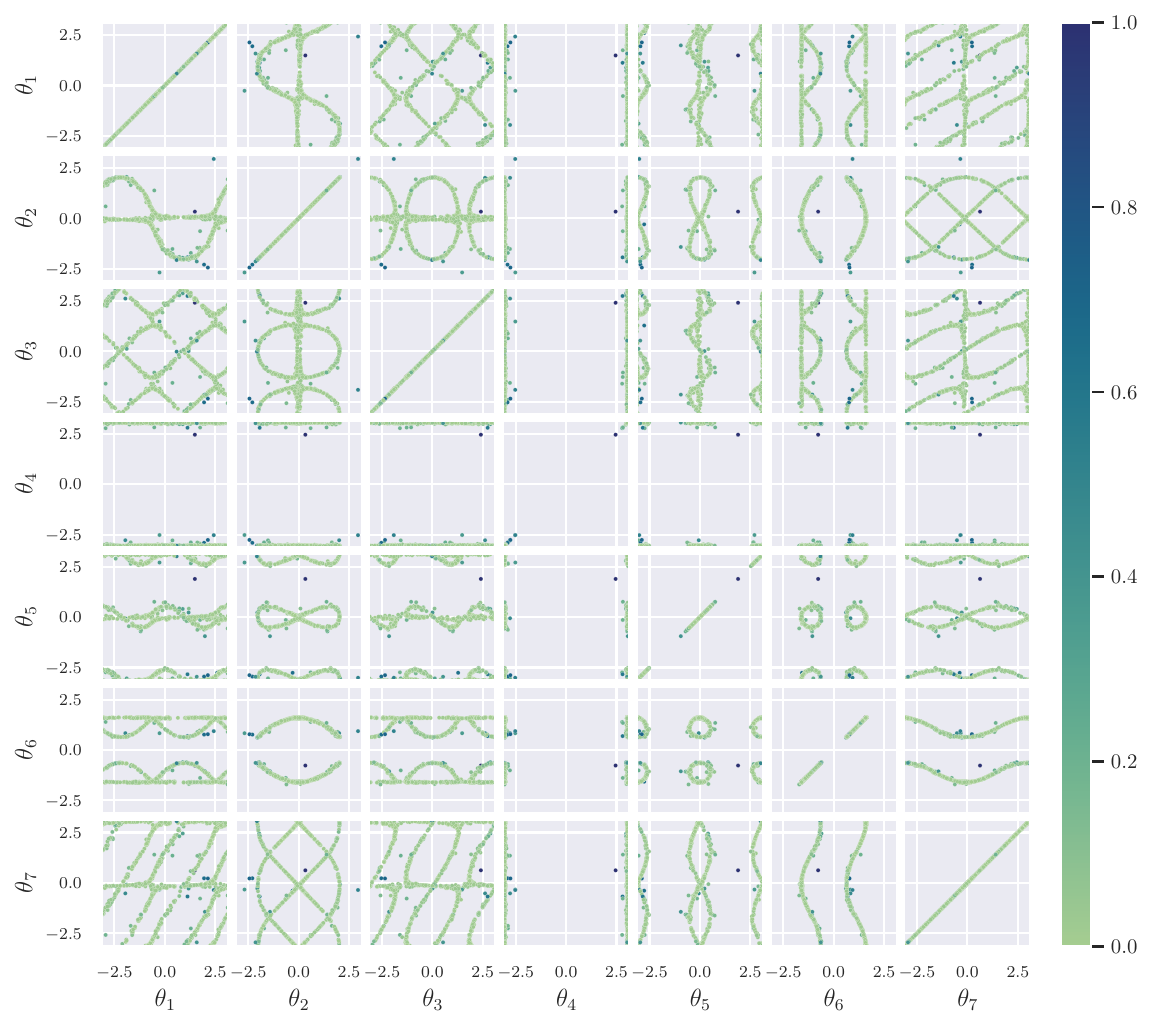}
  \caption{Pairwise plotting of 1,000 samples from GGIK (using a single model trained on all of the test manipulators) for a single Kuka goal pose. Each sub-plot contains samples on the 2-dimensional torus for joint angles $\theta_i$ and $\theta_j$, leading to a symmetrical pattern (i.e., the upper triangle is a transposition of the lower triangle). The hue of each sample is proportionate to the end-effector's pose error (the sum of the Euclidean distance in metres and the angular distance in radians). For this particular pose, the continuous and orderly curves produced by accurate solutions indicate that GGIK is able to learn a relatively complex distribution over a large, varied solution set. Samples with higher errors appear to happen sporadically.}\label{fig:kuka_pairplot}
\end{figure*}

\begin{figure*}[t]
  \centering
  \includegraphics[width=0.95\textwidth]{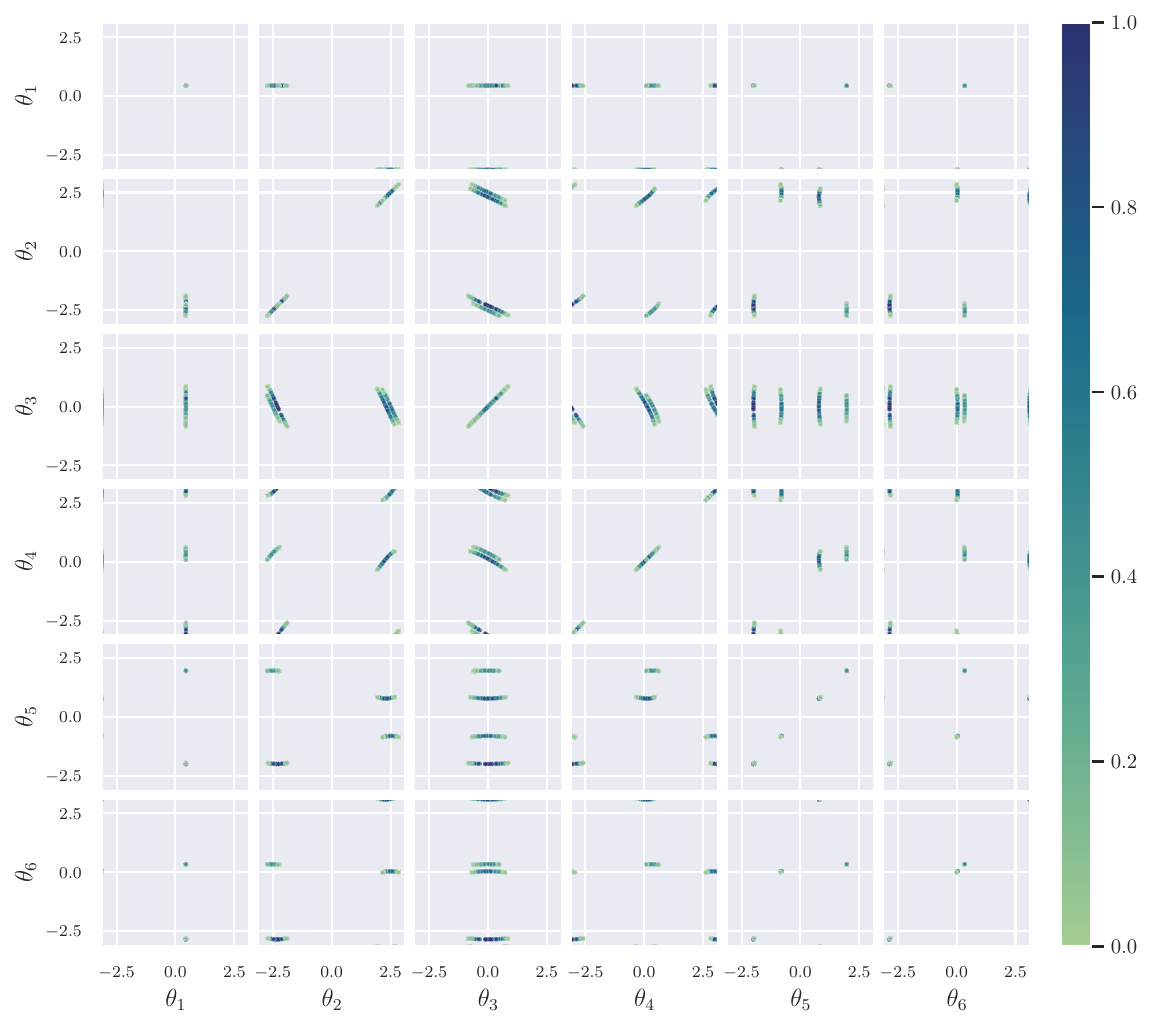}
  \caption{Pairwise plotting of 1,000 samples from GGIK (using a single model trained on all of the test manipulators) for a single UR10 goal pose. Each sub-plot contains samples on the 2-dimensional torus for joint angles $\theta_i$ and $\theta_j$, leading to a symmetrical pattern (i.e., the upper triangle is a transposition of the lower triangle). The hue of each sample is proportionate to the end-effector's pose error. For this particular goal pose, the less accurate samples from GGIK appear to be attempts at \emph{interpolating} between high accuracy discrete clusters (i.e., those with lighter colouring).}
  \label{fig:ur10_pairplot}
\end{figure*} 

In addition to providing accurate solutions, GGIK is able to uniformly sample an approximation of the set of feasible configurations for a given goal pose. 
We qualitatively demonstrate this capability by plotting pairs of joint angle variables for multiple solutions sampled from GGIK for a single query. 
An example of this visualization procedure for 1,000 samples for the Kuka arm is displayed in \cref{fig:kuka_pairplot}.
Each of the ${7 \choose 2} = 21$ plots in the symmetric upper and lower triangles of the array in \cref{fig:kuka_pairplot} contains the sampled joint angles for a pair $\theta_i$ and $\theta_j$.
Since each pair contains samples from the torus $S^1 \times S^1$, the $x-$ and $y-$ axes ``wrap around" at $\pi \equiv -\pi$.
The rendered hue of each sample is based on its pose error, which is defined as the sum of the Euclidean distance in metres and the angular distance in radians. \footnote{This ``informal" metric can be made into a physically commensurable quantity by multiplying the angular error by a scaling factor with units of metres per radian.}
Darker samples indicate less accurate solutions. 
The continuous and orderly curves produced by accurate solutions demonstrate that GGIK is able to learn a relatively complex distribution over a large, varied solution set.

Since the UR10 is a (non-redundant) 6-DOF manipulator, we expect the same visualization for this arm to consist of discrete clusters.
\cref{fig:ur10_pairplot} demonstrates that this is indeed the case, with an interesting phenomenon occurring for this particular goal pose: the less accurate samples from GGIK appear to be attempts at \emph{interpolating} between high accuracy clusters (i.e., those with lighter colouring).
This does not occur for all poses sampled, and a quantitative clustering experiment confirms that GGIK is in fact learning a few accurate clusters for most poses.

\begin{figure}
	\centering
	\begin{subfigure}[t]{\columnwidth}
  \centering
    \begin{adjustbox}{width=0.9\linewidth}
      	\includegraphics[width=\textwidth]{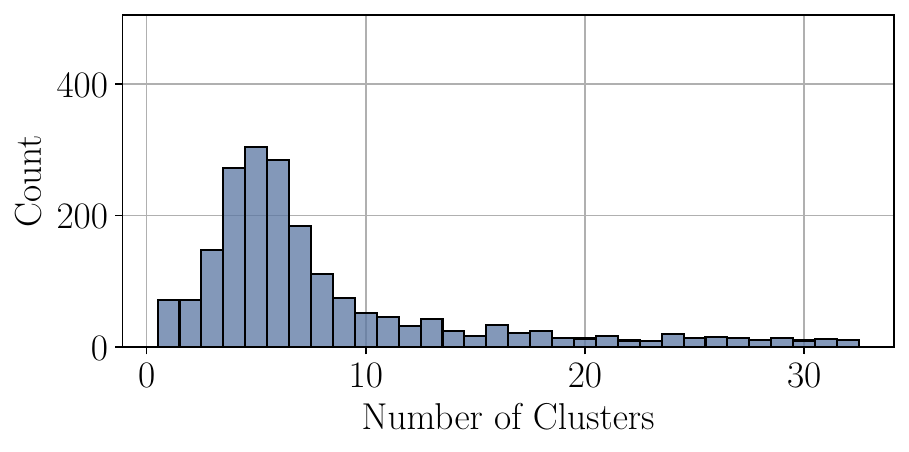}
    \end{adjustbox}
    	\vspace{-2mm}
		\caption{$\epsilon=0.01$}\label{fig:ur10_clusters_eps001}
    \vspace{2mm}
	\end{subfigure}
	\begin{subfigure}[t]{\columnwidth}
		\centering
		\begin{adjustbox}{width=0.9\linewidth}
      		\includegraphics[width=\textwidth]{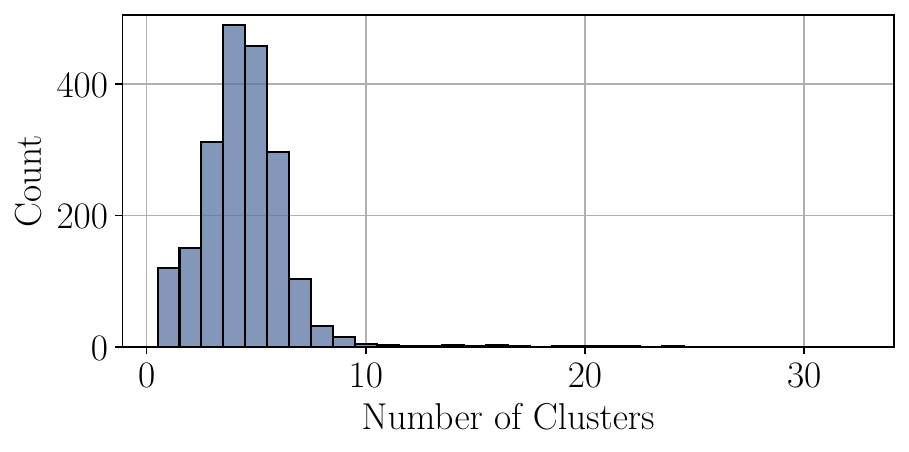}
		\end{adjustbox}
		\vspace{-2mm}
		\caption{$\epsilon=0.1$}\label{fig:ur10_clusters_eps01}
		\vspace{2mm}
	\end{subfigure}
	\begin{subfigure}[t]{\columnwidth}
		\centering
		\begin{adjustbox}{width=0.9\linewidth}
      		\includegraphics[width=\textwidth]{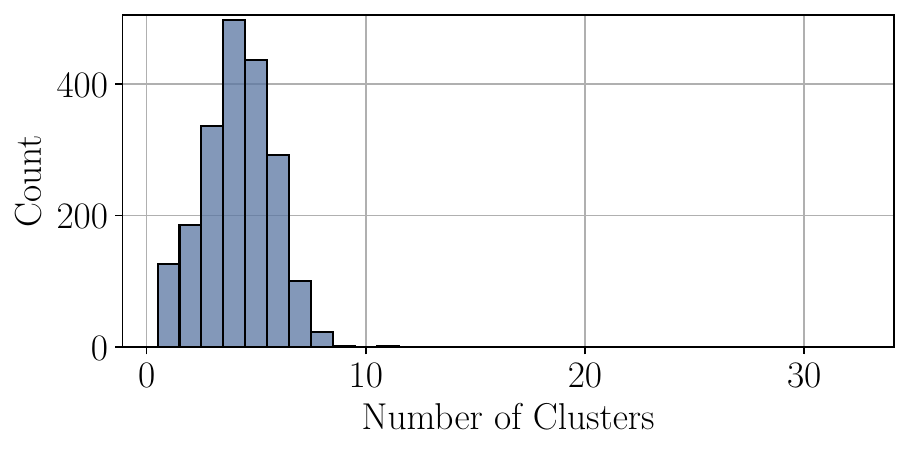}
		\end{adjustbox}
		\vspace{-2mm}
		\caption{$\epsilon=1$}\label{fig:ur10_clusters_eps1}
	\end{subfigure}
	\caption{Distribution of the number of clusters produced by GGIK for 2,000 goal poses. Each clustering was performed on 32 GGIK samples with the DBSCAN algorithm and the specified radius $\epsilon$. The $\epsilon$ value determines the sensitivity of the clustering procedure. The similarity between all three distributions indicate that the clustering procedure is fairly stable.}
	\label{fig:ur10_clusters}
\end{figure}

Using 32 GGIK samples for each of 2,000 random goal poses, we computed a density-based spatial clustering of applications with noise (DBSCAN) \cite{schubert2017dbscan} using the product metric of the chordal distance for $S^1$.
DBSCAN was selected because it supports data on topological manifolds and has very few parameters to tune.
\cref{fig:ur10_clusters} contains histograms displaying the distribution of the number of clusters recovered by DBSCAN with a variety of values for the radius parameter $\epsilon$. 
In all three cases, the minimum cluster size was set to one in order to allow for clusters containing a single sample.
Ideally, we would like to see the number of clusters concentrated around eight, which is the maximum number of solutions for the UR10.
In \cref{fig:ur10_clusters_eps001}, $\epsilon$ is small and noisy clusters are erroneously separated.
However, the similarity between the distributions in \cref{fig:ur10_clusters_eps01} and \cref{fig:ur10_clusters_eps1} indicates that this is a fairly stable clustering procedure.
Indeed, in both histograms there are very few poses for which GGIK produced greater than eight clusters, while the median number of clusters is four. 
We note that in the example shown in \cref{fig:ur10_pairplot}, the erroneous interpolation between accurate clusters reduces the amount of clusters from eight to four. 
This interpolation behaviour partially explains the reduced number of clusters found by DBSCAN in the samples produced by GGIK for the UR10.
Whether our approach can be modified to consistently produce the maximum number of solutions possible in the case of 6-DOF manipulators remains to be seen, but the results in \cref{fig:ur10_clusters} show that our current model does not learn distributions concentrated around a single solution.

\subsection{Time Taken to Generate Multiple Samples}
\label{subsec:expAppendix1}
\begin{figure}
  \centering
    \begin{adjustbox}{width=0.9\linewidth}
      	\includegraphics[width=\textwidth]{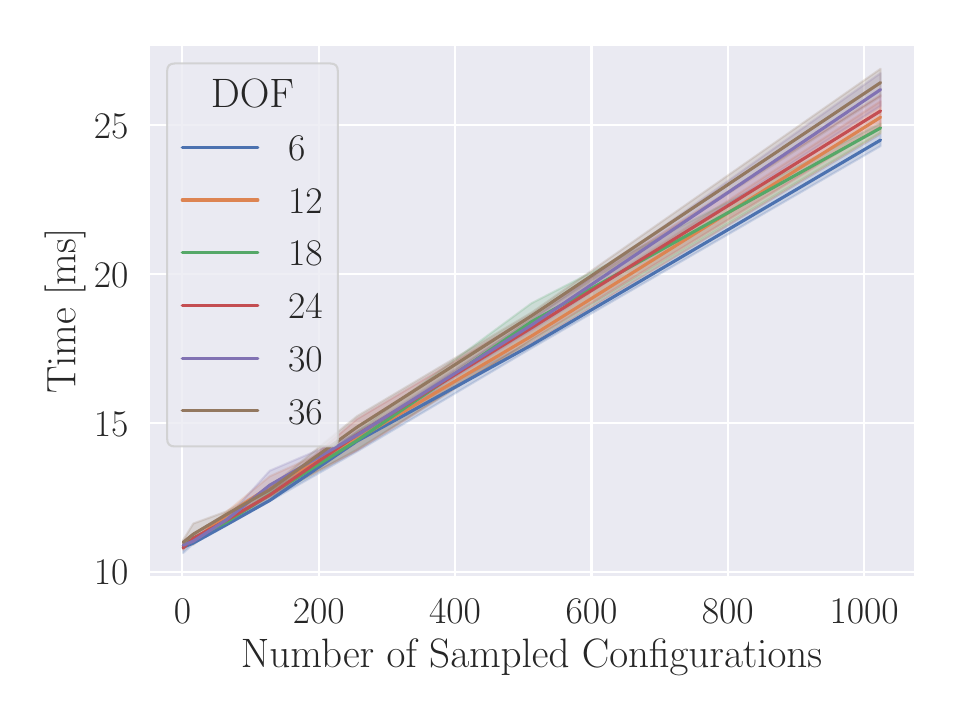}
    \end{adjustbox}
	\caption{Plot of the time taken in milliseconds as a function of the number of sampled solutions generated by GGIK for robots of varying degrees of freedoms (DOF). For a range of robots with 6- to 36-DOF, it takes approximately 10 milliseconds to generate about 10 samples, and approximately 25 milliseconds to generate 1,000 samples. Shaded region shows 95\% confidence interval. The size of the robot (i.e., size of the graph) and the number of samples do not significantly change the inference time due to efficient parallelization from the GPU.}
	\label{fig:timing_plot}
\end{figure}

We studied the scaling of the computation time of GGIK as a function of the number of solution generated and the number of DOF in the robot (i.e., the size of the robot).
GGIK is able to take advantage of the GPU architecture and yield very low per-solution solve times when amortized across larger batches.
This could be used, for example, in sampling-based motion planning and workspace analysis.
The results displayed in \cref{fig:timing_plot} demonstrate that GGIK was able to generate 10 samples in 10 milliseconds and 1,000 samples in 25 milliseconds for a range of robots with 6- to 36-DOF.
\subsection{GGIK and Numerical Solvers}
\label{subsec:exp2}

\begin{table*}
  \centering

\input{tables/experiment_3_local_new.tex}
  \caption{Comparison of GGIK (using a single model trained on all of the test manipulators) with TRAC-IK~\cite{beeson2015trac} on the task of producing 32 solutions for 10000 randomly-generated IK problems. TRAC-IK is initialized using configurations sampled randomly (TRAC-IK + rand) or from GGIK (TRAC-IK + GGIK). The error statistics are presented as the mean and mean minimum and maximum error per problem, as well as the two quartiles of the distribution. The computation time per solution is computed as the average time required for a single solution to be produced per problem.}\label{table:exp2_local}
\end{table*}

The distribution that GGIK learns can be efficiently sampled to produce multiple approximate solutions in parallel.
These samples can be further refined with relatively little additional computational cost by using them to initialize optimization-based methods.
\cref{table:exp2_local} shows the results of repeating the experiment in~\cref{subsec:exp1} using the TRAC-IK~\cite{beeson2015trac} inverse kinematics solver, where the results are averaged over all problems and all robots.

  In our study, we measured the total per-problem computation time for GGIK, including both the time taken to sample point sets and to retrieve associated joint angles on the GPU, operations which can be executed concurrently.
  We used a single GGIK model trained on all of the test manipulators.
  We then calculated the per-sample computation time by dividing the total computation time by the number of samples (32). 
  Specifically, GGIK alone achieved an average per-sample computation time of 0.34 ms, totalling approximately 11 ms per problem.
  For increased accuracy, these GGIK-generated samples were used to initialize the TRAC-IK solver.
  These initializations helped TRAC-IK converge to a solution in 0.16 ms on average, totalling 0.5 ms per-sample or 16 ms per problem when GGIK computation time is included.
  Conversely, randomly initialized TRAC-IK required on average 0.58 ms per sample, totalling 19.2 ms per problem.
  The notably higher convergence time for random initializations can be ascribed to the solver requiring a much larger number of steps to reach a feasible solution from a random point in the configuration space.

  The accuracy of both TRAC-IK instances, in terms of position and rotation errors, is superior to that of GGIK alone, reflecting the known advantage of local solvers over learning-based methods.
  However, as~\cref{subsec:multimod} illustrates, GGIK offers control over the diversity of solutions, sampling extensively across the solution space.
  %
  %
  In contrast, we cannot predict which solutions local solvers will converge to with random initializations.
  Additionally, using GGIK to initialize TRAC-IK produced solutions that were more accurate than a randomly initialized TRAC-IK.

  Our experiments, as detailed in~\cref{subsec:expAppendix1}, demonstrate that leveraging GPU parallelization allows the generation of a significantly larger number of samples with minimal additional computation time.
  This finding underscores the efficiency of combining GGIK with local solvers, particularly in complex inverse kinematics scenarios featuring multiple spatial constraints and joint limits.
  Overall, these results indicate that GGIK can be efficiently used in tandem with local solvers in order to produce a set of highly accurate and diverse solutions with lower or equal computation times.

\subsection{Ablation Study on the Equivariant Network Architecture}
\label{subsec:exp4}

\begin{table*}
  \centering
  \input{tables/experiment_4.tex}
  \caption{Comparison of different network architectures. EGNN outperforms existing architectures that are not equivariant in terms of overall accuracy and test ELBO. Dashed results are models with output point sets that were too far from a valid joint configuration and diverged during the configuration reconstruction procedure. Using 32 samples from the learned distribution per goal pose, the error statistics are presented as the mean and mean minimum and maximum error per problem, as well as the two quartiles of the error distribution.}
  \label{table:exp4}
\end{table*}

We conducted an ablation experiment to evaluate the importance of capturing the underlying $\LieGroupE{n}$ equivariance of graphical IK in our learning architecture.
We compared the accuracy of GGIK when using the EGNN network~\cite{satorras2021n} as opposed to four commonly used GNN layers that are not $\LieGroupE{n}$ equivariant: GRAPHsage~\cite{hamilton2017inductive}, GAT~\cite{velickovic2018graph}, GCN~\cite{Kipf2016tc} and MPNN~\cite{gilmer2017neural}.
We matched the number of parameters of each GNN architecture as closely as possible and kept all other experimental parameters fixed.
The dataset for this experiment was the same one used in~\cref{subsec:exp1}. 
For each GNN architecture, the results were averaged over all manipulators as shown in \cref{table:exp4}.
Out of the five different architectures that we compared, only the EGNN and MPNN outputted point sets that could be successfully mapped to valid joint configurations.
Point sets that were too far from those representing a valid joint configuration resulted in the configuration reconstruction procedure diverging.
The equivariant EGNN model outperformed all other models in terms of the overall solution accuracy and ELBO value attained on a held-out test set.
\subsection{Joint Limits}
\label{subsec:expjointlimits}

GGIK is able to learn a constrained distribution of solutions that obey joint limits. 
We demonstrated this by training a variant of GGIK using 512,000 samples from a Panda manipulator that are within specific joint limits. 
In our experiment with the Panda we used the joint limit values provided on the website of the manufacturer, Franka Emika. 

\begin{figure}
  \centering
    \begin{adjustbox}{width=1.0\linewidth}
      	\includegraphics[width=\textwidth]{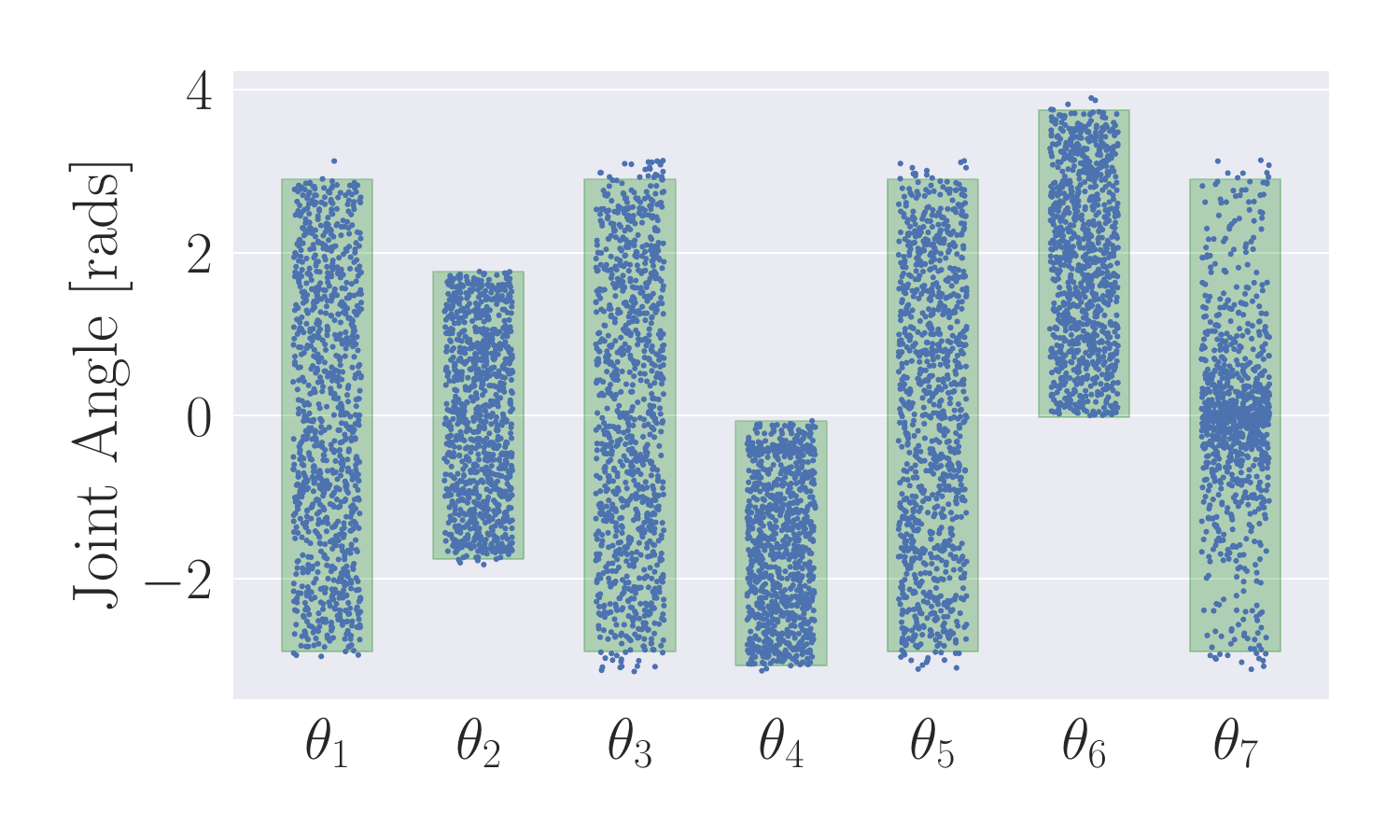}
    \end{adjustbox}
	\caption{Joint angles of solutions generated by GGIK for 1000 Panda IK problems. The green shaded bars denote joint limits of the samples used during training. The solutions are denoted in blue, where each column represents a specific joint. GGIK is able to learn a constrained distribution of solutions with samples that are almost always within the joint limits.}
	\label{fig:joint_limit_plot}
\end{figure}

At test time, we created 1,000 goal poses and used GGIK to solve for the associated joint angles.
In \cref{fig:joint_limit_plot}, we visualized the joint angles of the solutions sampled by GGIK.
The green shaded bars denote the range of valid joint angles.
The joint angle solutions are denoted in blue, where each column represents a specific joint of the Panda. 
Notably, we observe that the samples almost always obeyed joint limits.
A handful of joint solutions were slightly outside of the limits.
This is to be expected since solutions are sampled from a learned distribution and joint limits are not explicitly enforced. 
In practice, we could run a few iterations of a numerical solver to guarantee that the solutions are within the joint limits.
More generally, the findings illustrated in \cref{fig:joint_limit_plot} show that the learned distribution of solutions from GGIK can be influenced by curating the training data.

\section{Limitations}
\label{sec:limitations}
Our approach is not without its limitations.
GGIK outputs may require post-processing by local optimization methods in applications with extremely low pose error tolerances.
Moreover, the sampled solutions may require filtering in order to adhere to specific task or configuration space constraints.
While filtering for common constraints (e.g., joint limits, pose error, proximity, etc.) may be carried out in parallel over all samples, conditioning the learned distribution on such constraints would likely be preferable.

Another potential limitation of our approach is inherited from our chosen representation~\cite{2021_Maric_Riemannian_B}, where the partially-connected distance graphs representing the IK problem only uniquely represent robots whose neighbouring joints have coplanar rotation axes.
Unlike the matrix completion method used in~\cite{2021_Maric_Riemannian_B}, neural networks are capable of representing arbitrary functions based on data, and are therefore able to circumvent this ambiguity when learning the solution distribution for an individual robot with joint rotation axes that are not coplanar.
However, the generalization capability a single model jointly learning solution distributions for multiple robots with non coplanar rotation axes may be more challenging due to the aforementioned ambiguity.

Finally, the uniform sampling strategy used in this paper to generate training data may not work well for more complicated classes of robots such as humanoids or quadrupeds.
In these cases, not all joint solutions are equally desireable due to additional potential constraints such as maintaining upright posture and static equilibrium.
Furthermore, uniform sampling suffers from the curse of dimensionality as the number of robot DOF increases.
We believe that a single potential solution to both of these issues would be targeted and biased sampling (e.g., only sampling solutions that maximize manipulability or where the humanoid remains upright). 

\section{Conclusion and Future Work}
\label{sec:conclusion}

We have presented GGIK, a generative graphical IK solver that is able to produce multiple diverse and accurate solutions in parallel across many different manipulator types.
This capability is achieved through a distance-geometric representation of the IK problem in concert with GNNs and generative modelling.
The accuracy of the generated solutions points to the potential of GGIK as both a standalone solver and as an initialization method for local optimization methods.
To the best of our knowledge, this is the first learned IK model that can generate multiple solutions for robots not present in the training set.
Importantly, because GGIK is fully differentiable, it can be incorporated as a flexible IK component that is part of an end-to-end learning-based robotic manipulation framework.
GGIK provides a framework for learned general IK---a universal solver (or initializer) that can provide multiple diverse solutions for any manipulator structure in a way that complements or replaces numerical optimization.

As interesting future work, we would like to learn constrained distributions of robot configurations that account for obstacles in the task space; obstacles can be incorporated in the distance-geometric representation of IK~\cite{2021_Maric_Riemannian_B, 2022_Giamou_Convex} as nodes.
%
%
%
Learning an obstacle- and collision-aware distribution would yield a solver that implements collision avoidance by way of message passing between manipulator and obstacle nodes.
However, the exact representations of the manipulator and obstacle shapes would require further investigation.
For example, a sphere-based representation, as used in \cite{lembono2021learning}, would be relatively simple to implement but may be too conservative when compared to some other 3D neural representation that might contain more information and higher resolution.
It would also be interesting to investigate the idea of fine-tuning a generic model trained on multiple robots to perform better on a specific robot with only a few iterations of learning.
We observed that GGIK seems to sometimes erroneously interpolate between the discrete solution sets of the UR10, and investigating ways to mitigate this behaviour is a promising research direction. 
Finally, it would be interesting to extend GGIK for use with other classes of robots such as kinematic trees (e.g., quadrupeds and humanoids). In particular, we would like to consider the case where multiple goal poses are specified and to investigate the sampling strategies needed to efficiently generate proper training datasets.
\bibliographystyle{IEEEtran}
\bibliography{references}
\begin{IEEEbiography}	
[{\includegraphics[width=1in,height=1.25in,clip,keepaspectratio]{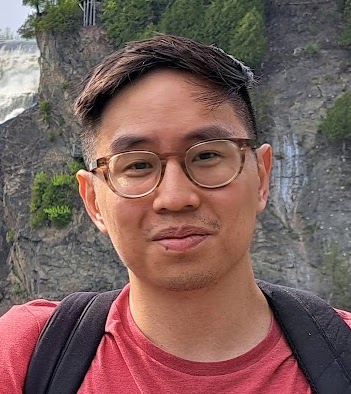}}]
{Oliver Limoyo} received his B.Eng. degree in mechanical engineering from McGill University in 2016. He is currently pursuing a Ph.D. degree at the Space and Terrestrial Autonomous Robotic Systems (STARS) laboratory at the University of Toronto. His research interests include the integration of generative models for robotics, reinforcement learning and imitation learning.
\end{IEEEbiography}

\begin{IEEEbiography}	[{\includegraphics[width=1in,height=1.25in,clip,keepaspectratio]{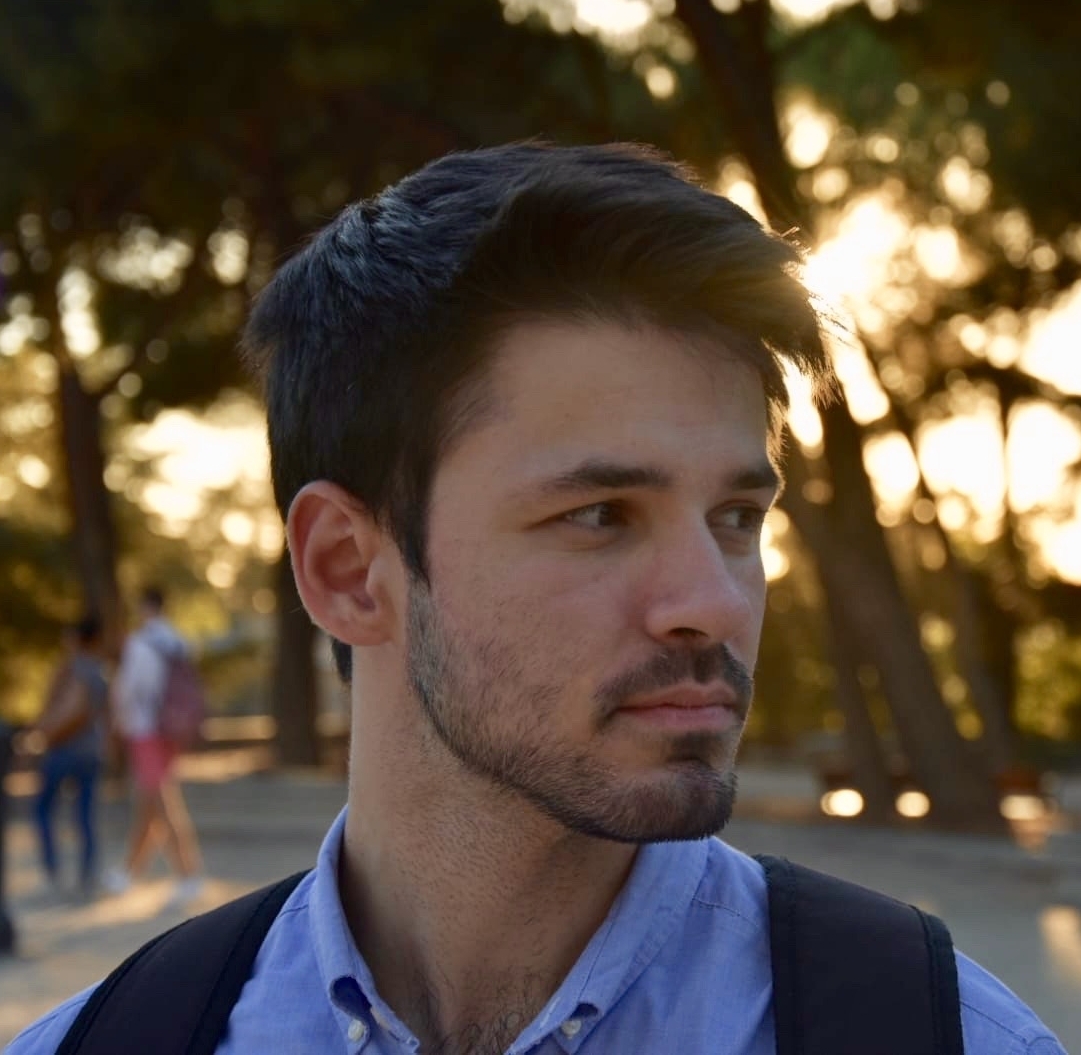}}]{Filip Mari\'c}
  Received his Ph.D.\ degree from the University of Toronto, Canada and University of Zagreb, Croatia in 2023.
  He received his B.Sc.\ and M.Sc.\ Degrees in electrical engineering and information technology from the University of Zagreb in 2015 and 2017, respectively.
  He is currently a senior research AI research scientist at the Samsung AI Center in Montreal, Canada. His research interests include robotic manipulation, planning and multi-modal deep learning.
\end{IEEEbiography}

\begin{IEEEbiography}[{\includegraphics[width=1in,height=1.25in,clip,keepaspectratio]{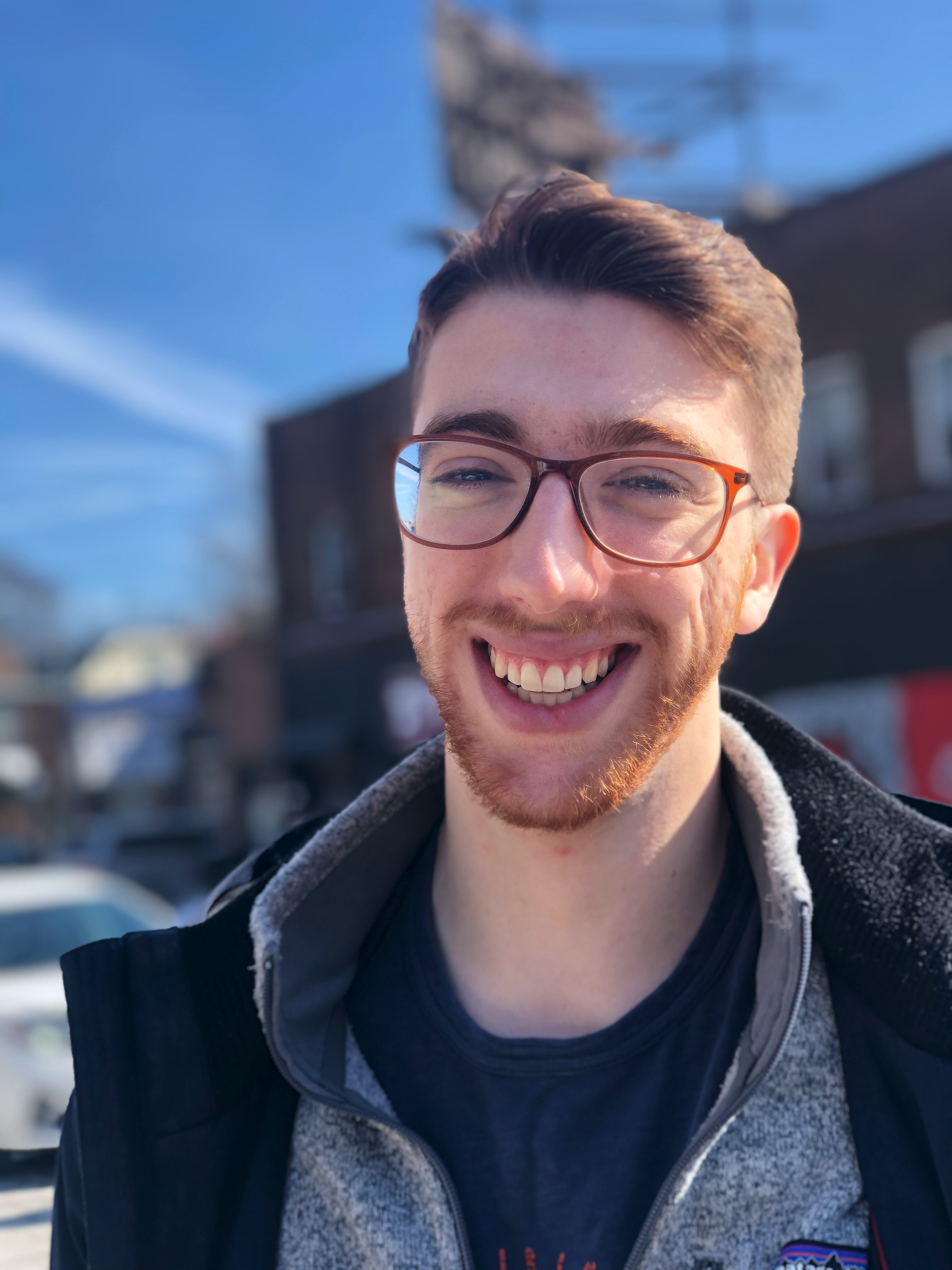}}]{Matthew Giamou}
	received his Ph.D.\ degree from the University of Toronto in 2023.
	He is currently a postdoctoral researcher at the Northeastern University Robust Autonomy Laboratory (NEURAL), where he is developing efficient convex optimization methods for large-scale robust robotic perception.
	His research interests also include multi-robot systems, sensor calibration, and the integration of learned and classical models for state estimation.
\end{IEEEbiography}

\begin{IEEEbiography}[{\includegraphics[width=1in,height=1.25in,clip,keepaspectratio]{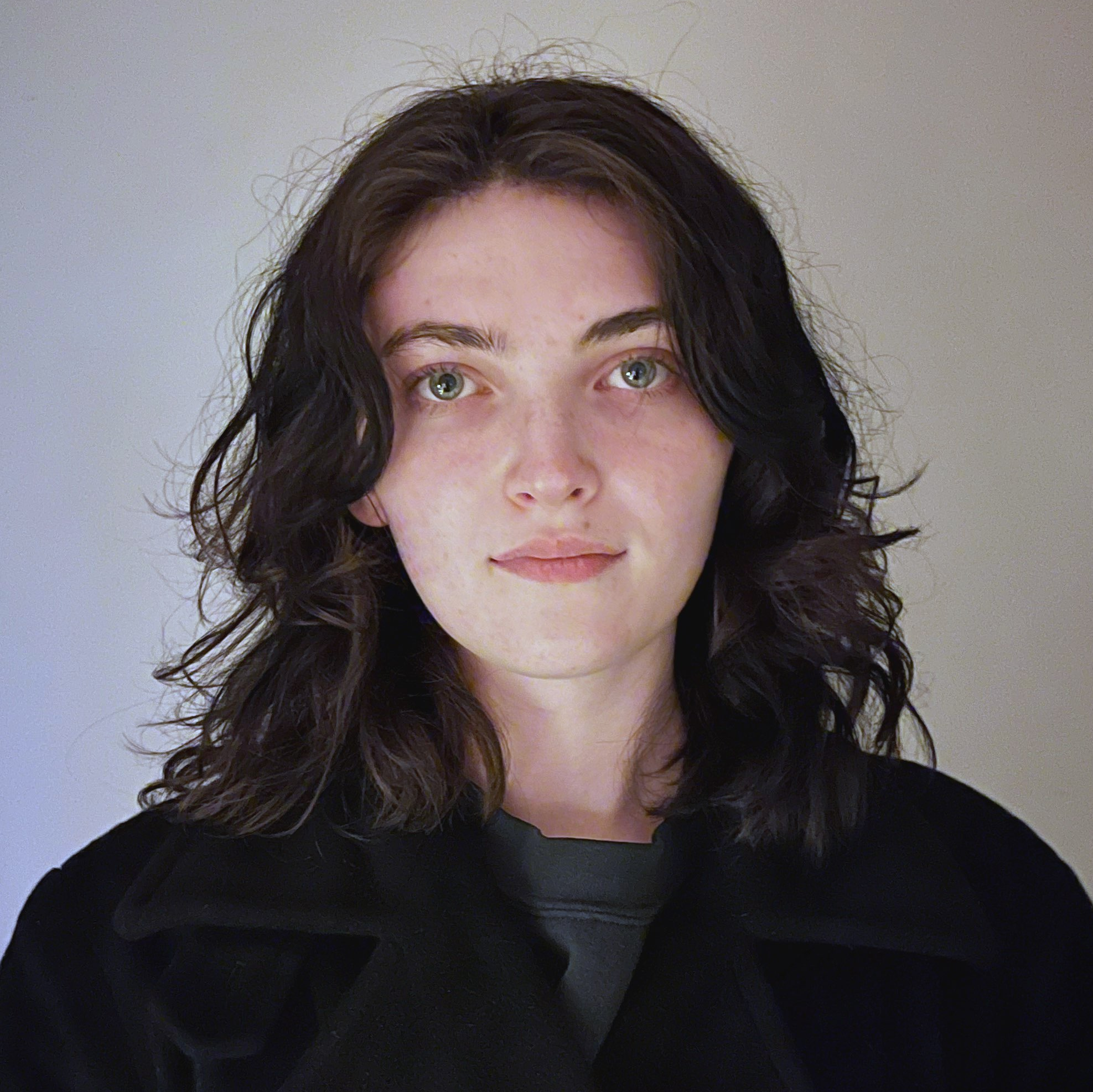}}]{Petra Alexson}
	is a fourth-year Engineering Science student at the University of Toronto, specializing in electrical and computer engineering. She was an undergraduate research assistant at the Space and Terrestrial Autonomous Robotic Systems (STARS) laboratory at the University of Toronto Institute for Aerospace Studies in the summer of 2021. During her time in STARS, she investigated several machine learning-based solutions to the inverse kinematics problem. She expects to receive her B.A.Sc in 2024.
\end{IEEEbiography}

\begin{IEEEbiography}[{\includegraphics[width=1in,height=1.25in,clip,keepaspectratio]{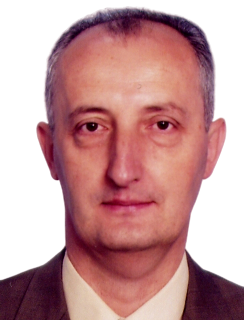}}]{Ivan Petrovi\'c} received the Master of Science and the Ph.D. degrees from FER Zagreb, Zagreb, Croatia, in 1990 and 1998, respectively. He is a Professor and the Head of the Laboratory for Autonomous Systems and Mobile Robotics, Faculty of Electrical Engineering and Computing, University of Zagreb, Zagreb, Croatia. He has published about 60 journal and 200 conference papers. His current research interest includes advanced control and estimation techniques and their application in autonomous systems and robotics. Prof. Petrovi\'c is a Full Member of the Croatian Academy of Engineering, and the Chair of the IFAC Technical Committee on Robotics.
\end{IEEEbiography}

\begin{IEEEbiography}[{\includegraphics[width=1in,height=1.25in,clip,keepaspectratio]{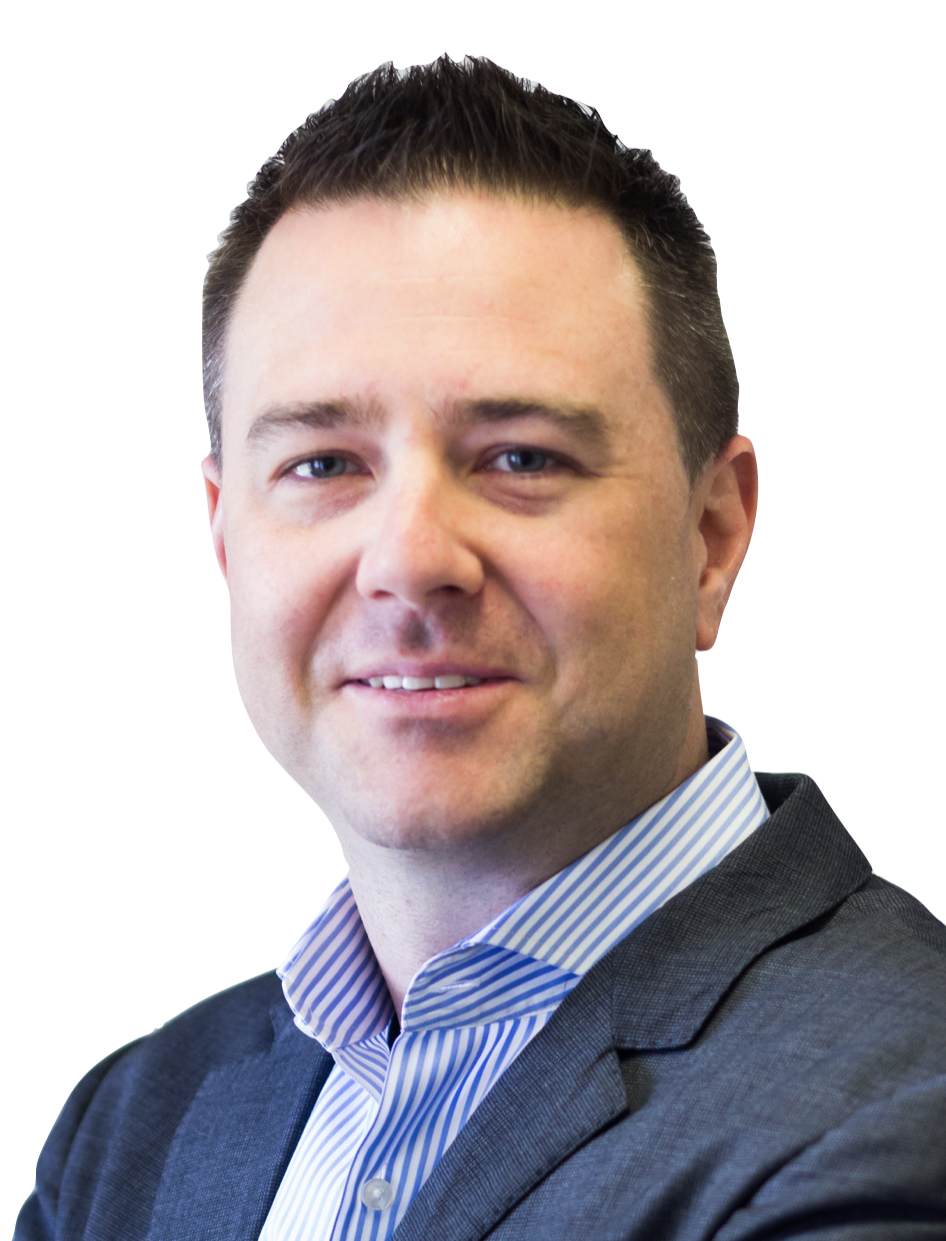}}]{Jonathan Kelly} received his Ph.D.\ degree from the University of Southern California, Los Angeles, USA, in 2011. From 2011 to 2013 he was a postdoctoral fellow in the Computer Science and Artificial Intelligence Laboratory at the Massachusetts Institute of Technology, Cambridge, USA. He is currently an associate professor and director of the Space and Terrestrial Autonomous Robotic Systems (STARS) Laboratory, University of Toronto Institute for Aerospace Studies, Toronto, Canada. Prof. Kelly holds the Tier II Canada Research Chair in Collaborative Robotics. His research interests include perception, planning, and learning for interactive robotic systems.
\end{IEEEbiography}
\end{document}

%% file: preamble.tex
\usepackage{amsmath}
\usepackage{amssymb}
\usepackage{amsthm} 

\usepackage{multicol}
\usepackage{float}
\makeatletter
\let\NAT@parse\undefined
\makeatother
\usepackage[colorlinks]{hyperref}
\usepackage{cite}

\usepackage{xparse}
\usepackage{amsfonts}
\usepackage{dsfont}
\usepackage{graphicx}
\usepackage{url}
\usepackage{booktabs}
\usepackage[para]{threeparttable}
\usepackage{multirow}

\usepackage{xcolor}
\usepackage{framed}
\usepackage{caption}
\usepackage{varwidth}
\usepackage{subcaption}
\usepackage[utf8]{inputenc}
\usepackage{pgfplots}
\usepackage{adjustbox}
\usepackage{tikz}
\usepackage{soul}
\DeclareUnicodeCharacter{2212}{−}
\usepgfplotslibrary{groupplots,dateplot}
\usetikzlibrary{patterns,shapes.arrows}
\pgfplotsset{compat=newest}
\usepackage{flushend}

\usepackage{pifont}

\usepackage[ruled,vlined]{algorithm2e}
\SetKwInOut{Parameters}{Parameters}

\SetKwComment{Comment}{$\triangleright$\ }{}
\SetCommentSty{mycommfont}
\usepackage{setspace}

\usepackage[capitalize]{cleveref}
\usepackage{xspace}
\usepackage[normalem]{ulem}

\colorlet{shadecolor}{gray!10}

\newtheorem*{DGP*}{Distance Geometry Problem}



%% file: notation.tex
\NewDocumentCommand\bbm{}{ \begin{bmatrix} }
\NewDocumentCommand\ebm{}{ \end{bmatrix} }
\NewDocumentCommand\Vector{m}{ \boldsymbol{\mathbf{#1}} }
\NewDocumentCommand\Matrix{m}{ \boldsymbol{\mathbf{#1}} }
\NewDocumentCommand\Transpose{m}{ \left.{#1}\right.^T }

\NewDocumentCommand\Norm{m}{\left\Vert#1\right\Vert }

\NewDocumentCommand\Cardinality{m}{ \left\vert#1\right\vert }
\NewDocumentCommand\Real{}{ \mathbb{R} }

\NewDocumentCommand\LieGroupSO{m}{ \mathrm{SO}(#1) }

\NewDocumentCommand\LieGroupE{m}{ \mathrm{E}(#1) }
\NewDocumentCommand\LieGroupSE{m}{ \mathrm{SE}(#1) }

\NewDocumentCommand\MatLog{m}{\mathrm{Log}\left({#1}\right)}

\NewDocumentCommand\AbsoluteValue{m}{ \Cardinality{m} }

\NewDocumentCommand\Transform{}{ \Matrix{T} }

\NewDocumentCommand\Defined{}{\triangleq}

\NewDocumentCommand\TaskSpace{}{\mathcal{T}}
\NewDocumentCommand\Task{}{\boldsymbol{\tau}}
\NewDocumentCommand\ConfigurationSpace{}{\mathcal{C}}
\NewDocumentCommand\Config{}{\boldsymbol{\theta}}

\newlength{\minuslength}

%% file: tables/experiment_2_individual.tex


\begin{tabular}{lrrrrrrrrrrrr}
\toprule
 Robot & \multicolumn{5}{c}{Err. Pos. [mm]} & \multicolumn{5}{c}{Err. Rot. [deg]}\\
 & mean & min & max & Q$_{1}$ & Q$_{3}$ & mean & min & max & Q$_{1}$ & Q$_{3}$ &  & \\
\midrule
KUKA & 4.6 & 2.4 & 7.2 & 3.7 & 5.5 & 0.4 & 0.2 & 0.5 & 0.3 & 0.4 \\
LWA4D & 4.3 & 1.3 & 8.1 & 3.0 & 5.4 & 0.4 & 0.1 & 0.6 & 0.3 & 0.5 \\
LWA4P & 3.8 & 1.6 & 6.5 & 2.9 & 4.7 & 0.3 & 0.1 & 0.4 & 0.2 & 0.4 \\
Panda & 5.8 & 1.6 & 11.1 & 4.0 & 7.4 & 0.4 & 0.1 & 0.7 & 0.3 & 0.5 \\
Panda with IKFlow \cite{ames2021ikflow} & 7.7 & - & - & - & - & 2.8 & - & - & - & - \\
Panda with IKNet \cite{2022_Bensadoun_Neural} & 31.0 & - & - & 13.5 & 48.6 & - & - & - & - & - \\
UR10 & 5.5 & 2.7 & 8.6 & 4.4 & 6.6 & 0.3 & 0.1 & 0.5 & 0.2 & 0.4 \\
UR10 with DT \cite{vonoehsen_comparison_2020} & 43.5 & 0.7 & 1124.7 & 16.2 & 45.9 & 15.3 & 0.2 & 177.7 & 5.7 & 18.3 \\
12-DOF & 15.3 & 4.6 & 25.4 & 11.5 & 19.2 & 0.7 & 0.1 & 1.3 & 0.5 & 1.0 \\
\bottomrule
\end{tabular}

%% file: tables/experiment_3.tex
\begin{tabular}{lrrrrrrrrrrr}
\toprule
 & \multicolumn{5}{c}{Err. Pos. [mm]} & \multicolumn{5}{c}{Err. Rot. [deg]}\\
 & mean & min & max & Q$_{1}$ & Q$_{3}$ & mean & min & max & Q$_{1}$ & Q$_{3}$ &  \\
Robot &  &  &  &  &  &  &  &  &  &  &  \\
\midrule
KUKA & 4.9 & 2.4 & 7.8 & 3.9 & 5.9 & 0.4 & 0.2 & 0.6 & 0.3 & 0.5 \\
LWA4D & 4.8 & 2.1 & 7.9 & 3.7 & 5.8 & 0.4 & 0.2 & 0.6 & 0.4 & 0.5 \\
LWA4P & 6.2 & 3.5 & 9.0 & 5.2 & 7.1 & 0.5 & 0.3 & 0.6 & 0.4 & 0.5 \\
Panda & 6.3 & 2.8 & 10.7 & 4.9 & 7.7 & 0.5 & 0.2 & 0.7 & 0.4 & 0.6 \\
UR10 & 9.3 & 6.9 & 11.5 & 8.6 & 10.0 & 0.5 & 0.4 & 0.7 & 0.5 & 0.6 \\
\bottomrule
\end{tabular}

%% file: figures/tikz/box_comp_pos.tex
\definecolor{color0}{rgb}{0.917647058823529,0.917647058823529,0.949019607843137}
\definecolor{color1}{rgb}{0.347058823529412,0.458823529411765,0.641176470588235}
\definecolor{color2}{rgb}{0.798529411764706,0.536764705882353,0.389705882352941}

\begin{axis}[
axis background/.style={fill=color0},
axis line style={white},
legend cell align={left},
legend style={
  fill opacity=0.8,
  draw opacity=1,
  text opacity=1,
  at={(0.03,0.97)},
  anchor=north west,
  draw=white!80!black,
  fill=color0
},
x grid style={white},
xmajorticks=false,
xmajorticks=true,
xmin=-0.5, xmax=4.5,
xtick pos=left,
xtick style={color=white!15!black},
xtick={0,1,2,3,4},
xticklabels={KUKA,LWA4D,LWA4P,Panda,UR10},
y grid style={white},
ylabel={Err. Position [mm]},
ymajorgrids,
ymajorticks=false,
ymajorticks=true,
ymin=0.195167177915573, ymax=32.7548477709293,
ytick pos=left,
ytick style={color=white!15!black}
]
\path [draw=white!29.8039215686275!black, fill=color1, semithick]
(axis cs:-0.396,5.49974119663239)
--(axis cs:-0.004,5.49974119663239)
--(axis cs:-0.004,8.20346283912659)
--(axis cs:-0.396,8.20346283912659)
--(axis cs:-0.396,5.49974119663239)
--cycle;
\path [draw=white!29.8039215686275!black, fill=color2, semithick]
(axis cs:0.004,5.21281182765961)
--(axis cs:0.396,5.21281182765961)
--(axis cs:0.396,8.46486210823059)
--(axis cs:0.004,8.46486210823059)
--(axis cs:0.004,5.21281182765961)
--cycle;
\path [draw=white!29.8039215686275!black, fill=color1, semithick]
(axis cs:0.604,5.54477310180664)
--(axis cs:0.996,5.54477310180664)
--(axis cs:0.996,7.89215087890625)
--(axis cs:0.604,7.89215087890625)
--(axis cs:0.604,5.54477310180664)
--cycle;
\path [draw=white!29.8039215686275!black, fill=color2, semithick]
(axis cs:1.004,4.38121020793915)
--(axis cs:1.396,4.38121020793915)
--(axis cs:1.396,6.90575098991394)
--(axis cs:1.004,6.90575098991394)
--(axis cs:1.004,4.38121020793915)
--cycle;
\path [draw=white!29.8039215686275!black, fill=color1, semithick]
(axis cs:1.604,4.93947410583496)
--(axis cs:1.996,4.93947410583496)
--(axis cs:1.996,7.98148787021637)
--(axis cs:1.604,7.98148787021637)
--(axis cs:1.604,4.93947410583496)
--cycle;
\path [draw=white!29.8039215686275!black, fill=color2, semithick]
(axis cs:2.004,4.43676567077637)
--(axis cs:2.396,4.43676567077637)
--(axis cs:2.396,7.32093715667725)
--(axis cs:2.004,7.32093715667725)
--(axis cs:2.004,4.43676567077637)
--cycle;
\path [draw=white!29.8039215686275!black, fill=color1, semithick]
(axis cs:2.604,12.8012583255768)
--(axis cs:2.996,12.8012583255768)
--(axis cs:2.996,20.1962084770203)
--(axis cs:2.604,20.1962084770203)
--(axis cs:2.604,12.8012583255768)
--cycle;
\path [draw=white!29.8039215686275!black, fill=color2, semithick]
(axis cs:3.004,9.81258678436279)
--(axis cs:3.396,9.81258678436279)
--(axis cs:3.396,14.2093186378479)
--(axis cs:3.004,14.2093186378479)
--(axis cs:3.004,9.81258678436279)
--cycle;
\path [draw=white!29.8039215686275!black, fill=color1, semithick]
(axis cs:3.604,6.4149295091629)
--(axis cs:3.996,6.4149295091629)
--(axis cs:3.996,12.4747769832611)
--(axis cs:3.604,12.4747769832611)
--(axis cs:3.604,6.4149295091629)
--cycle;
\path [draw=white!29.8039215686275!black, fill=color2, semithick]
(axis cs:4.004,7.09434306621552)
--(axis cs:4.396,7.09434306621552)
--(axis cs:4.396,12.0789425373077)
--(axis cs:4.004,12.0789425373077)
--(axis cs:4.004,7.09434306621552)
--cycle;
\draw[draw=white!29.8039215686275!black,fill=color1,line width=0.3pt] (axis cs:0,0) rectangle (axis cs:0,0);
\addlegendimage{ybar,ybar legend,draw=black,fill=color1};
\addlegendentry{$512,000$};

\draw[draw=white!29.8039215686275!black,fill=color2,line width=0.3pt] (axis cs:0,0) rectangle (axis cs:0,0);
\addlegendimage{ybar,ybar legend,draw=black,fill=color2};
\addlegendentry{$2,560,000$};

\addplot [semithick, white!29.8039215686275!black, forget plot]
table {%
-0.2 5.49974119663239
-0.2 2.53846764564514
};
\addplot [semithick, white!29.8039215686275!black, forget plot]
table {%
-0.2 8.20346283912659
-0.2 12.2501201629639
};
\addplot [semithick, white!29.8039215686275!black, forget plot]
table {%
-0.298 2.53846764564514
-0.102 2.53846764564514
};
\addplot [semithick, white!29.8039215686275!black, forget plot]
table {%
-0.298 12.2501201629639
-0.102 12.2501201629639
};
\addplot [semithick, white!29.8039215686275!black, forget plot]
table {%
0.2 5.21281182765961
0.2 1.90102350711823
};
\addplot [semithick, white!29.8039215686275!black, forget plot]
table {%
0.2 8.46486210823059
0.2 13.3262271881104
};
\addplot [semithick, white!29.8039215686275!black, forget plot]
table {%
0.102 1.90102350711823
0.298 1.90102350711823
};
\addplot [semithick, white!29.8039215686275!black, forget plot]
table {%
0.102 13.3262271881104
0.298 13.3262271881104
};
\addplot [semithick, white!29.8039215686275!black, forget plot]
table {%
0.8 5.54477310180664
0.8 2.65603470802307
};
\addplot [semithick, white!29.8039215686275!black, forget plot]
table {%
0.8 7.89215087890625
0.8 11.4070158004761
};
\addplot [semithick, white!29.8039215686275!black, forget plot]
table {%
0.702 2.65603470802307
0.898 2.65603470802307
};
\addplot [semithick, white!29.8039215686275!black, forget plot]
table {%
0.702 11.4070158004761
0.898 11.4070158004761
};
\addplot [semithick, white!29.8039215686275!black, forget plot]
table {%
1.2 4.38121020793915
1.2 1.6751526594162
};
\addplot [semithick, white!29.8039215686275!black, forget plot]
table {%
1.2 6.90575098991394
1.2 10.6414804458618
};
\addplot [semithick, white!29.8039215686275!black, forget plot]
table {%
1.102 1.6751526594162
1.298 1.6751526594162
};
\addplot [semithick, white!29.8039215686275!black, forget plot]
table {%
1.102 10.6414804458618
1.298 10.6414804458618
};
\addplot [semithick, white!29.8039215686275!black, forget plot]
table {%
1.8 4.93947410583496
1.8 1.79389953613281
};
\addplot [semithick, white!29.8039215686275!black, forget plot]
table {%
1.8 7.98148787021637
1.8 12.5333223342896
};
\addplot [semithick, white!29.8039215686275!black, forget plot]
table {%
1.702 1.79389953613281
1.898 1.79389953613281
};
\addplot [semithick, white!29.8039215686275!black, forget plot]
table {%
1.702 12.5333223342896
1.898 12.5333223342896
};
\addplot [semithick, white!29.8039215686275!black, forget plot]
table {%
2.2 4.43676567077637
2.2 1.79233610630035
};
\addplot [semithick, white!29.8039215686275!black, forget plot]
table {%
2.2 7.32093715667725
2.2 11.6444568634033
};
\addplot [semithick, white!29.8039215686275!black, forget plot]
table {%
2.102 1.79233610630035
2.298 1.79233610630035
};
\addplot [semithick, white!29.8039215686275!black, forget plot]
table {%
2.102 11.6444568634033
2.298 11.6444568634033
};
\addplot [semithick, white!29.8039215686275!black, forget plot]
table {%
2.8 12.8012583255768
2.8 5.87155771255493
};
\addplot [semithick, white!29.8039215686275!black, forget plot]
table {%
2.8 20.1962084770203
2.8 31.2748622894287
};
\addplot [semithick, white!29.8039215686275!black, forget plot]
table {%
2.702 5.87155771255493
2.898 5.87155771255493
};
\addplot [semithick, white!29.8039215686275!black, forget plot]
table {%
2.702 31.2748622894287
2.898 31.2748622894287
};
\addplot [semithick, white!29.8039215686275!black, forget plot]
table {%
3.2 9.81258678436279
3.2 5.45934629440308
};
\addplot [semithick, white!29.8039215686275!black, forget plot]
table {%
3.2 14.2093186378479
3.2 20.7949924468994
};
\addplot [semithick, white!29.8039215686275!black, forget plot]
table {%
3.102 5.45934629440308
3.298 5.45934629440308
};
\addplot [semithick, white!29.8039215686275!black, forget plot]
table {%
3.102 20.7949924468994
3.298 20.7949924468994
};
\addplot [semithick, white!29.8039215686275!black, forget plot]
table {%
3.8 6.4149295091629
3.8 2.05880618095398
};
\addplot [semithick, white!29.8039215686275!black, forget plot]
table {%
3.8 12.4747769832611
3.8 21.5557231903076
};
\addplot [semithick, white!29.8039215686275!black, forget plot]
table {%
3.702 2.05880618095398
3.898 2.05880618095398
};
\addplot [semithick, white!29.8039215686275!black, forget plot]
table {%
3.702 21.5557231903076
3.898 21.5557231903076
};
\addplot [semithick, white!29.8039215686275!black, forget plot]
table {%
4.2 7.09434306621552
4.2 2.10084676742554
};
\addplot [semithick, white!29.8039215686275!black, forget plot]
table {%
4.2 12.0789425373077
4.2 19.537769317627
};
\addplot [semithick, white!29.8039215686275!black, forget plot]
table {%
4.102 2.10084676742554
4.298 2.10084676742554
};
\addplot [semithick, white!29.8039215686275!black, forget plot]
table {%
4.102 19.537769317627
4.298 19.537769317627
};
\addplot [semithick, white!29.8039215686275!black, forget plot]
table {%
-0.396 6.69059228897095
-0.004 6.69059228897095
};
\addplot [semithick, white!29.8039215686275!black, forget plot]
table {%
0.004 6.82561039924622
0.396 6.82561039924622
};
\addplot [semithick, white!29.8039215686275!black, forget plot]
table {%
0.604 6.55113387107849
0.996 6.55113387107849
};
\addplot [semithick, white!29.8039215686275!black, forget plot]
table {%
1.004 5.4935610294342
1.396 5.4935610294342
};
\addplot [semithick, white!29.8039215686275!black, forget plot]
table {%
1.604 6.21275639533997
1.996 6.21275639533997
};
\addplot [semithick, white!29.8039215686275!black, forget plot]
table {%
2.004 5.63547801971436
2.396 5.63547801971436
};
\addplot [semithick, white!29.8039215686275!black, forget plot]
table {%
2.604 15.8733048439026
2.996 15.8733048439026
};
\addplot [semithick, white!29.8039215686275!black, forget plot]
table {%
3.004 11.4952545166016
3.396 11.4952545166016
};
\addplot [semithick, white!29.8039215686275!black, forget plot]
table {%
3.604 8.635573387146
3.996 8.635573387146
};
\addplot [semithick, white!29.8039215686275!black, forget plot]
table {%
4.004 9.009108543396
4.396 9.009108543396
};
\end{axis}

%% file: figures/tikz/box_comp_rot.tex
\definecolor{color0}{rgb}{0.917647058823529,0.917647058823529,0.949019607843137}
\definecolor{color1}{rgb}{0.347058823529412,0.458823529411765,0.641176470588235}
\definecolor{color2}{rgb}{0.798529411764706,0.536764705882353,0.389705882352941}

\begin{axis}[
axis background/.style={fill=color0},
axis line style={white},
legend cell align={left},
legend style={
  fill opacity=0.8,
  draw opacity=1,
  text opacity=1,
  at={(0.03,0.97)},
  anchor=north west,
  draw=white!80!black,
  fill=color0
},
x grid style={white},
xmajorticks=false,
xmajorticks=true,
xmin=-0.5, xmax=4.5,
xtick pos=left,
xtick style={color=white!15!black},
xtick={0,1,2,3,4},
xticklabels={KUKA,LWA4D,LWA4P,Panda,UR10},
y grid style={white},
ylabel={Err. Rotation [\(\displaystyle \circ\)]},
ymajorgrids,
ymajorticks=false,
ymajorticks=true,
ymin=-0.00978304408490659, ymax=2.72502534799278,
ytick pos=left,
ytick style={color=white!15!black}
]
\path [draw=white!29.8039215686275!black, fill=color1, semithick]
(axis cs:-0.396,0.398230277001858)
--(axis cs:-0.004,0.398230277001858)
--(axis cs:-0.004,0.633519604802132)
--(axis cs:-0.396,0.633519604802132)
--(axis cs:-0.396,0.398230277001858)
--cycle;
\path [draw=white!29.8039215686275!black, fill=color2, semithick]
(axis cs:0.004,0.321373656392097)
--(axis cs:0.396,0.321373656392097)
--(axis cs:0.396,0.473338477313519)
--(axis cs:0.004,0.473338477313519)
--(axis cs:0.004,0.321373656392097)
--cycle;
\path [draw=white!29.8039215686275!black, fill=color1, semithick]
(axis cs:0.604,0.423186123371124)
--(axis cs:0.996,0.423186123371124)
--(axis cs:0.996,0.673710212111473)
--(axis cs:0.604,0.673710212111473)
--(axis cs:0.604,0.423186123371124)
--cycle;
\path [draw=white!29.8039215686275!black, fill=color2, semithick]
(axis cs:1.004,0.309012740850449)
--(axis cs:1.396,0.309012740850449)
--(axis cs:1.396,0.463182382285595)
--(axis cs:1.004,0.463182382285595)
--(axis cs:1.004,0.309012740850449)
--cycle;
\path [draw=white!29.8039215686275!black, fill=color1, semithick]
(axis cs:1.604,0.395271092653275)
--(axis cs:1.996,0.395271092653275)
--(axis cs:1.996,0.76219442486763)
--(axis cs:1.604,0.76219442486763)
--(axis cs:1.604,0.395271092653275)
--cycle;
\path [draw=white!29.8039215686275!black, fill=color2, semithick]
(axis cs:2.004,0.306795284152031)
--(axis cs:2.396,0.306795284152031)
--(axis cs:2.396,0.534527093172073)
--(axis cs:2.004,0.534527093172073)
--(axis cs:2.004,0.306795284152031)
--cycle;
\path [draw=white!29.8039215686275!black, fill=color1, semithick]
(axis cs:2.604,0.991580054163933)
--(axis cs:2.996,0.991580054163933)
--(axis cs:2.996,1.63808816671371)
--(axis cs:2.604,1.63808816671371)
--(axis cs:2.604,0.991580054163933)
--cycle;
\path [draw=white!29.8039215686275!black, fill=color2, semithick]
(axis cs:3.004,0.70172755420208)
--(axis cs:3.396,0.70172755420208)
--(axis cs:3.396,1.16079795360565)
--(axis cs:3.004,1.16079795360565)
--(axis cs:3.004,0.70172755420208)
--cycle;
\path [draw=white!29.8039215686275!black, fill=color1, semithick]
(axis cs:3.604,0.377356693148613)
--(axis cs:3.996,0.377356693148613)
--(axis cs:3.996,0.796511322259903)
--(axis cs:3.604,0.796511322259903)
--(axis cs:3.604,0.377356693148613)
--cycle;
\path [draw=white!29.8039215686275!black, fill=color2, semithick]
(axis cs:4.004,0.403619728982449)
--(axis cs:4.396,0.403619728982449)
--(axis cs:4.396,0.720062807202339)
--(axis cs:4.004,0.720062807202339)
--(axis cs:4.004,0.403619728982449)
--cycle;
\draw[draw=white!29.8039215686275!black,fill=color1,line width=0.3pt] (axis cs:0,0) rectangle (axis cs:0,0);

\draw[draw=white!29.8039215686275!black,fill=color2,line width=0.3pt] (axis cs:0,0) rectangle (axis cs:0,0);

\addplot [semithick, white!29.8039215686275!black, forget plot]
table {%
-0.2 0.398230277001858
-0.2 0.193370565772057
};
\addplot [semithick, white!29.8039215686275!black, forget plot]
table {%
-0.2 0.633519604802132
-0.2 0.975549221038818
};
\addplot [semithick, white!29.8039215686275!black, forget plot]
table {%
-0.298 0.193370565772057
-0.102 0.193370565772057
};
\addplot [semithick, white!29.8039215686275!black, forget plot]
table {%
-0.298 0.975549221038818
-0.102 0.975549221038818
};
\addplot [semithick, white!29.8039215686275!black, forget plot]
table {%
0.2 0.321373656392097
0.2 0.176751792430878
};
\addplot [semithick, white!29.8039215686275!black, forget plot]
table {%
0.2 0.473338477313519
0.2 0.699152588844299
};
\addplot [semithick, white!29.8039215686275!black, forget plot]
table {%
0.102 0.176751792430878
0.298 0.176751792430878
};
\addplot [semithick, white!29.8039215686275!black, forget plot]
table {%
0.102 0.699152588844299
0.298 0.699152588844299
};
\addplot [semithick, white!29.8039215686275!black, forget plot]
table {%
0.8 0.423186123371124
0.8 0.189388796687126
};
\addplot [semithick, white!29.8039215686275!black, forget plot]
table {%
0.8 0.673710212111473
0.8 1.04632925987244
};
\addplot [semithick, white!29.8039215686275!black, forget plot]
table {%
0.702 0.189388796687126
0.898 0.189388796687126
};
\addplot [semithick, white!29.8039215686275!black, forget plot]
table {%
0.702 1.04632925987244
0.898 1.04632925987244
};
\addplot [semithick, white!29.8039215686275!black, forget plot]
table {%
1.2 0.309012740850449
1.2 0.169384777545929
};
\addplot [semithick, white!29.8039215686275!black, forget plot]
table {%
1.2 0.463182382285595
1.2 0.693995356559753
};
\addplot [semithick, white!29.8039215686275!black, forget plot]
table {%
1.102 0.169384777545929
1.298 0.169384777545929
};
\addplot [semithick, white!29.8039215686275!black, forget plot]
table {%
1.102 0.693995356559753
1.298 0.693995356559753
};
\addplot [semithick, white!29.8039215686275!black, forget plot]
table {%
1.8 0.395271092653275
1.8 0.153325036168098
};
\addplot [semithick, white!29.8039215686275!black, forget plot]
table {%
1.8 0.76219442486763
1.8 1.30538058280945
};
\addplot [semithick, white!29.8039215686275!black, forget plot]
table {%
1.702 0.153325036168098
1.898 0.153325036168098
};
\addplot [semithick, white!29.8039215686275!black, forget plot]
table {%
1.702 1.30538058280945
1.898 1.30538058280945
};
\addplot [semithick, white!29.8039215686275!black, forget plot]
table {%
2.2 0.306795284152031
2.2 0.114526428282261
};
\addplot [semithick, white!29.8039215686275!black, forget plot]
table {%
2.2 0.534527093172073
2.2 0.875623762607574
};
\addplot [semithick, white!29.8039215686275!black, forget plot]
table {%
2.102 0.114526428282261
2.298 0.114526428282261
};
\addplot [semithick, white!29.8039215686275!black, forget plot]
table {%
2.102 0.875623762607574
2.298 0.875623762607574
};
\addplot [semithick, white!29.8039215686275!black, forget plot]
table {%
2.8 0.991580054163933
2.8 0.339185684919357
};
\addplot [semithick, white!29.8039215686275!black, forget plot]
table {%
2.8 1.63808816671371
2.8 2.60071587562561
};
\addplot [semithick, white!29.8039215686275!black, forget plot]
table {%
2.702 0.339185684919357
2.898 0.339185684919357
};
\addplot [semithick, white!29.8039215686275!black, forget plot]
table {%
2.702 2.60071587562561
2.898 2.60071587562561
};
\addplot [semithick, white!29.8039215686275!black, forget plot]
table {%
3.2 0.70172755420208
3.2 0.306448072195053
};
\addplot [semithick, white!29.8039215686275!black, forget plot]
table {%
3.2 1.16079795360565
3.2 1.84028160572052
};
\addplot [semithick, white!29.8039215686275!black, forget plot]
table {%
3.102 0.306448072195053
3.298 0.306448072195053
};
\addplot [semithick, white!29.8039215686275!black, forget plot]
table {%
3.102 1.84028160572052
3.298 1.84028160572052
};
\addplot [semithick, white!29.8039215686275!black, forget plot]
table {%
3.8 0.377356693148613
3.8 0.153035759925842
};
\addplot [semithick, white!29.8039215686275!black, forget plot]
table {%
3.8 0.796511322259903
3.8 1.4234756231308
};
\addplot [semithick, white!29.8039215686275!black, forget plot]
table {%
3.702 0.153035759925842
3.898 0.153035759925842
};
\addplot [semithick, white!29.8039215686275!black, forget plot]
table {%
3.702 1.4234756231308
3.898 1.4234756231308
};
\addplot [semithick, white!29.8039215686275!black, forget plot]
table {%
4.2 0.403619728982449
4.2 0.144001603126526
};
\addplot [semithick, white!29.8039215686275!black, forget plot]
table {%
4.2 0.720062807202339
4.2 1.19112753868103
};
\addplot [semithick, white!29.8039215686275!black, forget plot]
table {%
4.102 0.144001603126526
4.298 0.144001603126526
};
\addplot [semithick, white!29.8039215686275!black, forget plot]
table {%
4.102 1.19112753868103
4.298 1.19112753868103
};
\addplot [semithick, white!29.8039215686275!black, forget plot]
table {%
-0.396 0.493611663579941
-0.004 0.493611663579941
};
\addplot [semithick, white!29.8039215686275!black, forget plot]
table {%
0.004 0.384678170084953
0.396 0.384678170084953
};
\addplot [semithick, white!29.8039215686275!black, forget plot]
table {%
0.604 0.534731388092041
0.996 0.534731388092041
};
\addplot [semithick, white!29.8039215686275!black, forget plot]
table {%
1.004 0.371542826294899
1.396 0.371542826294899
};
\addplot [semithick, white!29.8039215686275!black, forget plot]
table {%
1.604 0.539346665143967
1.996 0.539346665143967
};
\addplot [semithick, white!29.8039215686275!black, forget plot]
table {%
2.004 0.398277938365936
2.396 0.398277938365936
};
\addplot [semithick, white!29.8039215686275!black, forget plot]
table {%
2.604 1.27237546443939
2.996 1.27237546443939
};
\addplot [semithick, white!29.8039215686275!black, forget plot]
table {%
3.004 0.88628813624382
3.396 0.88628813624382
};
\addplot [semithick, white!29.8039215686275!black, forget plot]
table {%
3.604 0.540557324886322
3.996 0.540557324886322
};
\addplot [semithick, white!29.8039215686275!black, forget plot]
table {%
4.004 0.539787203073502
4.396 0.539787203073502
};
\end{axis}

%% file: tables/experiment_x.tex
\begin{tabular}{lrrrrrrrrrrr}
\toprule
Robot & \multicolumn{5}{c}{Err. Pos. [mm]} & \multicolumn{5}{c}{Err. Rot. [deg]} \\
 & mean & min & max & Q$_{1}$ & Q$_{3}$ & mean & min & max & Q$_{1}$ & Q$_{3}$ &  \\
\midrule
KUKA & 39.7 & 11.4 & 75.5 & 27.8 & 50.4 & 2.5 & 0.5 & 4.6 & 1.7 & 3.3 \\
Lwa4d & 35.7 & 10.2 & 68.3 & 25.0 & 45.5 & 2.3 & 0.5 & 4.3 & 1.6 & 3.1 \\
Lwa4p & 22.5 & 7.3 & 40.4 & 16.3 & 28.2 & 1.5 & 0.4 & 2.6 & 1.0 & 1.9 \\
UR10 & 46.5 & 22.1 & 69.7 & 37.9 & 55.3 & 2.0 & 0.6 & 3.5 & 1.4 & 2.5 \\
\bottomrule
\end{tabular}

%% file: tables/experiment_3_local_new.tex
\begin{tabular}{lrrrrrrrrrrrrr}
\toprule
 & \multicolumn{5}{c}{Err. Pos. [mm]} & \multicolumn{5}{c}{Err. Rot. [deg]} & \multicolumn{2}{c}{Soln. Time [ms]} \\
 & mean & min & max & Q$_{1}$ & Q$_{3}$ & mean & min & max & Q$_{1}$ & Q$_{3}$ & mean & std &  \\
Method &  &  &  &  &  &  &  &  &  &  &  &  &  \\
\midrule
GGIK & 7.09 & 2.52 & 12.63 & 5.16 & 8.84 & 0.52 & 0.15 & 0.84 & 0.38 & 0.66 & 0.34 & 0.03 \\
TracIK + GGIK & 0.15 & 0.01 & 0.86 & 0.04 & 0.18 & 0.01 & 0.00 & 0.08 & 0.00 & 0.02 & 0.50 & 0.09 \\
TracIK + rand & 0.17 & 0.00 & 0.97 & 0.02 & 0.22 & 0.02 & 0.00 & 0.07 & 0.00 & 0.02 & 0.58 & 0.39 \\
\bottomrule
\end{tabular}

%% file: tables/experiment_4.tex
\begin{tabular}{lrrrrrrrrrrrrr}
\toprule
 Model Name & \multicolumn{5}{c}{Err. Pos. [mm]} & \multicolumn{5}{c}{Err. Rot. [deg]} & Test ELBO \\
 & mean & min & max & Q$_1$ & Q$_3$ & mean & min & max & Q$_1$ & Q$_3$ \\
\midrule
EGNN~\cite{satorras2021n} & 4.6 & 1.5 & 8.5 & 3.3 & 5.8 & 0.4 & 0.1 & 0.6 & 0.3 & 0.4 & -0.05 \\
MPNN~\cite{gilmer2017neural} & 143.2 & 62.9 & 273.7 & 113.1 & 169.1 & 17.7 & 5.3 & 13.6 & 21.6 & 34.1 & -8.3 \\
GAT~\cite{velickovic2018graph} & - & - & - & - & - & - & - & - & - & - & -12.41 \\
GCN~\cite{Kipf2016tc} & - & - & - & - & - & - & - & - & - & - & -12.42 \\
GRAPHsage~\cite{hamilton2017inductive} & - & - & - & - & - & - & - & - & - & - & -10.5 \\
\bottomrule
\end{tabular}

%% file: manuscript.bbl
\begin{thebibliography}{10}
\providecommand{\url}[1]{#1}
\csname url@samestyle\endcsname
\providecommand{\newblock}{\relax}
\providecommand{\bibinfo}[2]{#2}
\providecommand{\BIBentrySTDinterwordspacing}{\spaceskip=0pt\relax}
\providecommand{\BIBentryALTinterwordstretchfactor}{4}
\providecommand{\BIBentryALTinterwordspacing}{\spaceskip=\fontdimen2\font plus
\BIBentryALTinterwordstretchfactor\fontdimen3\font minus
  \fontdimen4\font\relax}
\providecommand{\BIBforeignlanguage}[2]{{%
\expandafter\ifx\csname l@#1\endcsname\relax
\typeout{** WARNING: IEEEtran.bst: No hyphenation pattern has been}%
\typeout{** loaded for the language `#1'. Using the pattern for}%
\typeout{** the default language instead.}%
\else
\language=\csname l@#1\endcsname
\fi
#2}}
\providecommand{\BIBdecl}{\relax}
\BIBdecl

\bibitem{vonoehsen_comparison_2020}
T.~{von Oehsen}, A.~Fabisch, S.~Kumar, and F.~Kirchner, ``Comparison of
  {{Distal Teacher Learning}} with {{Numerical}} and {{Analytical Methods}} to
  {{Solve Inverse Kinematics}} for {{Rigid}}-{{Body Mechanisms}},''
  \emph{arXiv:2003.00225 [cs]}, 2020.

\bibitem{aristidouInverseKinematicsTechniques2018}
A.~Aristidou, J.~Lasenby, Y.~Chrysanthou, and A.~Shamir, ``Inverse kinematics
  techniques in computer graphics: A survey,'' \emph{Computer Graphics Forum},
  vol.~37, no.~6, pp. 35--58, 2018.

\bibitem{ren2020learning}
H.~Ren and P.~Ben-Tzvi, ``Learning inverse kinematics and dynamics of a robotic
  manipulator using generative adversarial networks,'' \emph{Robotics and
  Autonomous Systems}, vol. 124, p. 103386, 2020.

\bibitem{ho2022selective}
C.-K. Ho and C.-T. King, ``Selective inverse kinematics: A novel approach to
  finding multiple solutions fast for high-dof robotic,'' \emph{arXiv preprint
  arXiv:2202.07869}, 2022.

\bibitem{ames2021ikflow}
B.~Ames, J.~Morgan, and G.~Konidaris, ``Ikflow: Generating diverse inverse
  kinematics solutions,'' \emph{{IEEE} Robotics and Automation Letters},
  vol.~7, no.~3, pp. 7177--7184, 2022.

\bibitem{lembono2021learning}
T.~S. Lembono, E.~Pignat, J.~Jankowski, and S.~Calinon, ``Learning constrained
  distributions of robot configurations with generative adversarial network,''
  \emph{IEEE Robotics and Automation Letters}, vol.~6, no.~2, pp. 4233--4240,
  2021.

\bibitem{beeson2015trac}
P.~Beeson and B.~Ames, ``{TRAC-IK}: An open-source library for improved solving
  of generic inverse kinematics,'' in \emph{Intl. Conf. on Humanoid Robots
  (Humanoids)}, 2015.

\bibitem{porta_branch-and-prune_2005}
J.~Porta, L.~Ros, F.~Thomas, and C.~Torras, ``A branch-and-prune solver for
  distance constraints,'' \emph{IEEE Trans. Robot.}, vol.~21, pp. 176--187,
  Apr. 2005.

\bibitem{2021_Maric_Riemannian_B}
F.~Maric, M.~Giamou, A.~W. Hall, S.~Khoubyarian, I.~Petrovic, and J.~Kelly,
  ``Riemannian optimization for distance-geometric inverse kinematics,''
  \emph{{IEEE} Transactions on Robotics}, 2021.

\bibitem{lee1988new}
H.-Y. Lee and C.-G. Liang, ``A new vector theory for the analysis of spatial
  mechanisms,'' \emph{Mech. Mach. Theory}, vol.~23, no.~3, pp. 209--217, 1988.

\bibitem{manocha1994efficient}
D.~Manocha and J.~F. Canny, ``Efficient inverse kinematics for general 6r
  manipulators,'' \emph{IEEE Trans. Robot.}

\bibitem{husty2007new}
M.~Husty, M.~Pfurner, and H.~Schr{\"o}cker, ``A new and efficient algorithm for
  the inverse kinematics of a general serial 6r manipulator,'' \emph{Mech.
  Mach. Theory}, vol.~42, no.~1, pp. 66--81, 2007.

\bibitem{diankov2010automated}
R.~Diankov, ``Automated construction of robotic manipulation programs,'' Ph.D.
  dissertation, Carnegie Mellon University, Pittsburgh, PA, September 2010.

\bibitem{whitney1969resolved}
D.~E. Whitney, ``Resolved motion rate control of manipulators and human
  prostheses,'' \emph{IEEE Transactions on man-machine systems}, vol.~10,
  no.~2, pp. 47--53, 1969.

\bibitem{sciavicco1986coordinate}
L.~Sciavicco and B.~Siciliano, ``Coordinate transformation: A solution
  algorithm for one class of robots,'' \emph{IEEE Transactions on Systems, Man
  and Cybernetics}, vol.~16, pp. 550 -- 559, 08 1986.

\bibitem{nakamura1987task}
Y.~Nakamura, H.~Hanafusa, and T.~Yoshikawa, ``Task-priority based redundancy
  control of robot manipulators,'' \emph{Int. J. Rob. Res.}, vol.~6, no.~2, pp.
  3--15, 1987.

\bibitem{lynch2017modern}
K.~M. Lynch and F.~C. Park, \emph{Modern robotics}.\hskip 1em plus 0.5em minus
  0.4em\relax Cambridge University Press, 2017.

\bibitem{boyd2004convex}
S.~Boyd and L.~Vandenberghe, \emph{Convex Optimization}.\hskip 1em plus 0.5em
  minus 0.4em\relax Cambridge University Press, 2004.

\bibitem{goldfarb1983numerically}
D.~Goldfarb and A.~Idnani, ``A numerically stable dual method for solving
  strictly convex quadratic programs,'' \emph{Mathematical programming},
  vol.~27, no.~1, pp. 1--33, 1983.

\bibitem{wachter2006implementation}
A.~W{\"a}chter and L.~T. Biegler, ``On the implementation of an interior-point
  filter line-search algorithm for large-scale nonlinear programming,''
  \emph{Mathematical programming}, vol. 106, pp. 25--57, 2006.

\bibitem{bambade2022prox}
A.~Bambade, S.~El-Kazdadi, A.~Taylor, and J.~Carpentier, ``Prox-qp: Yet another
  quadratic programming solver for robotics and beyond,'' in \emph{Robotics:
  Science and Systems (RSS)}, 2022.

\bibitem{siciliano2010robotics}
B.~Siciliano, L.~Sciavicco, L.~Villani, and G.~Oriolo, \emph{Robotics:
  Modelling, Planning and Control}.\hskip 1em plus 0.5em minus 0.4em\relax
  Springer Science \& Business Media, 2010.

\bibitem{manipulation}
\BIBentryALTinterwordspacing
R.~Tedrake, \emph{Robotic Manipulation}, 2022. [Online]. Available:
  \url{http://manipulation.mit.edu}
\BIBentrySTDinterwordspacing

\bibitem{engell2009projected}
M.~Engell-N{\o}rreg{\aa}rd and K.~Erleben, ``A projected non-linear conjugate
  gradient method for interactive inverse kinematics,'' in \emph{Intl. Conf. on
  Mathematical Modelling (MATHMOD)}, 2009.

\bibitem{deo1993adaptive}
A.~S. Deo and I.~D. Walker, ``Adaptive non-linear least squares for inverse
  kinematics,'' in \emph{IEEE Intl. Conf. on Robotics and Automation}, 1993.

\bibitem{erleben_solving_2019}
K.~Erleben and S.~Andrews, ``\BIBforeignlanguage{en}{Solving inverse kinematics
  using exact {{Hessian}} matrices},'' \emph{\BIBforeignlanguage{en}{Comput.
  Graph.}}, vol.~78, pp. 1--11, Feb. 2019.

\bibitem{nocedal1999numerical}
J.~Nocedal and S.~J. Wright, \emph{Numerical optimization}.\hskip 1em plus
  0.5em minus 0.4em\relax Springer, 1999.

\bibitem{kdl-url}
R.~Smits, ``{KDL}: {K}inematics and {D}ynamics {L}ibrary,''
  \url{http://www.orocos.org/kdl}.

\bibitem{kenwright2012inverse}
B.~Kenwright, ``Inverse kinematics--cyclic coordinate descent ({CCD}),''
  \emph{J. Graphics Tools}, vol.~16, no.~4, pp. 177--217, 2012.

\bibitem{Aristidou_2011}
A.~Aristidou and J.~Lasenby, ``{FABRIK}: A fast, iterative solver for the
  inverse kinematics problem,'' \emph{Graph. Models}, vol.~73, no.~5, p.
  243–260, Sep. 2011.

\bibitem{dejalonTwentyfiveYearsNatural2007}
J.~G. {de Jal{\'o}n}, ``\BIBforeignlanguage{en}{Twenty-five years of natural
  coordinates},'' \emph{\BIBforeignlanguage{en}{Multibody Sys. Dyn.}}, vol.~18,
  no.~1, pp. 15--33, Aug. 2007.

\bibitem{dai_global_2019}
H.~Dai, G.~Izatt, and R.~Tedrake, ``\BIBforeignlanguage{en}{Global inverse
  kinematics via mixed-integer convex optimization},''
  \emph{\BIBforeignlanguage{en}{Int. J. Rob. Res.}}, vol.~38, no. 12-13, pp.
  1420--1441, Oct. 2019.

\bibitem{yenamandra_convex_2019}
T.~Yenamandra, F.~Bernard, J.~Wang, F.~Mueller, and C.~Theobalt,
  ``\BIBforeignlanguage{en}{Convex {{Optimisation}} for {{Inverse
  Kinematics}}},'' \emph{\BIBforeignlanguage{en}{Intl. Conf. on 3D Vision
  (3DV)}}, pp. 318--327, 2019.

\bibitem{le2019kinematics}
T.~Le~Naour, N.~Courty, and S.~Gibet, ``Kinematics in the metric space,''
  \emph{Comput. Graph.}, vol.~84, pp. 13--23, 2019.

\bibitem{2022_Giamou_Convex}
M.~Giamou, F.~Maric, D.~M. Rosen, V.~Peretroukhin, N.~Roy, I.~Petrovic, and
  J.~Kelly, ``Convex iteration for distance-geometric inverse kinematics,''
  \emph{{IEEE} Robotics and Automation Letters}, 2022.

\bibitem{jordan1992forward}
M.~I. Jordan and D.~E. Rumelhart, ``Forward models: Supervised learning with a
  distal teacher,'' \emph{Cognitive science}, vol.~16, no.~3, pp. 307--354,
  1992.

\bibitem{bullock1993self}
D.~Bullock, S.~Grossberg, and F.~H. Guenther, ``A self-organizing neural model
  of motor equivalent reaching and tool use by a multijoint arm,''
  \emph{Journal of Cognitive Neuroscience}, vol.~5, no.~4, pp. 408--435, 1993.

\bibitem{bocsi2011learning}
B.~B{\'o}csi, D.~Nguyen-Tuong, L.~Csat{\'o}, B.~Schoelkopf, and J.~Peters,
  ``Learning inverse kinematics with structured prediction,'' in \emph{IEEE/RSJ
  Intl. Conf. on Intelligent Robots and Systems}.\hskip 1em plus 0.5em minus
  0.4em\relax IEEE, 2011.

\bibitem{villegasNeuralKinematicNetworks2018}
R.~Villegas, J.~Yang, D.~Ceylan, and H.~Lee, ``Neural kinematic networks for
  unsupervised motion retargetting,'' in \emph{IEEE Conf. on Computer Vision
  and Pattern Recognition (CVPR)}, June 2018.

\bibitem{ardizzone2018analyzing}
L.~Ardizzone, J.~Kruse, C.~Rother, and U.~K{\"o}the, ``Analyzing inverse
  problems with invertible neural networks,'' in \emph{Intl. Conf. on Learning
  Representations (ICLR)}, 2018.

\bibitem{kruse2021benchmarking}
J.~Kruse, L.~Ardizzone, C.~Rother, and U.~K{\"o}the, ``Benchmarking invertible
  architectures on inverse problems,'' \emph{ICML Workshop on Invertible Neural
  Networks, Normalizing Flows, and Explicit Likelihood Models}, 2021.

\bibitem{2022_Bensadoun_Neural}
R.~Bensadoun, S.~Gur, N.~Blau, and L.~Wolf, ``Neural inverse kinematic,'' in
  \emph{Intl. Conf. on Machine Learning (ICML)}, vol. 162, 2022.

\bibitem{ichnowskiDeepLearningGrasp}
J.~Ichnowski, Y.~Avigal, V.~Satish, and K.~Goldberg, ``Deep learning can
  accelerate grasp-optimized motion planning,'' \emph{Science Robotics},
  vol.~5, no.~48, p. eabd7710, 2020.

\bibitem{khan2020graph}
A.~Khan, A.~Ribeiro, V.~Kumar, and A.~G. Francis, ``Graph neural networks for
  motion planning,'' \emph{arXiv preprint arXiv:2006.06248}, 2020.

\bibitem{ichter2018learning}
B.~Ichter, J.~Harrison, and M.~Pavone, ``Learning sampling distributions for
  robot motion planning,'' in \emph{IEEE Intl. Conf. on Robotics and Automation
  (ICRA)}, 2018.

\bibitem{qureshi2020motion}
A.~H. Qureshi, Y.~Miao, A.~Simeonov, and M.~C. Yip, ``Motion planning networks:
  Bridging the gap between learning-based and classical motion planners,''
  \emph{IEEE Transactions on Robotics}, vol.~37, no.~1, pp. 48--66, 2020.

\bibitem{10161569}
J.~Urain, N.~Funk, J.~Peters, and G.~Chalvatzaki, ``Se(3)-diffusionfields:
  Learning smooth cost functions for joint grasp and motion optimization
  through diffusion,'' in \emph{IEEE Intl. Conf. on Robotics and Automation
  (ICRA)}, 2023.

\bibitem{GNNBook-ch11-liao}
R.~Liao, ``Graph neural networks: Graph generation,'' in \emph{Graph Neural
  Networks: Foundations, Frontiers, and Applications}, L.~Wu, P.~Cui, J.~Pei,
  and L.~Zhao, Eds.\hskip 1em plus 0.5em minus 0.4em\relax Singapore: Springer
  Singapore, 2022, pp. 225--250.

\bibitem{kipf2016variational}
T.~N. Kipf and M.~Welling, ``Variational graph auto-encoders,'' \emph{NeurIPS
  Bayesian Deep Learning Workshop}, 2016.

\bibitem{de2018molgan}
N.~De~Cao and T.~Kipf, ``{MolGAN: An implicit generative model for small
  molecular graphs},'' \emph{ICML Workshop on Theoretical Foundations and
  Applications of Deep Generative Models}, 2018.

\bibitem{li2018learning}
Y.~Li, O.~Vinyals, C.~Dyer, R.~Pascanu, and P.~Battaglia, ``Learning deep
  generative models of graphs,'' \emph{arXiv preprint arXiv:1803.03324}, 2018.

\bibitem{simm2019generative}
G.~N.~C. Simm and J.~M. Hern\'{a}ndez-Lobato, ``A generative model for
  molecular distance geometry,'' in \emph{Intl. Conf. on Machine Learning
  (ICML)}, 2020.

\bibitem{murray2017mathematical}
R.~M. Murray, Z.~Li, and S.~S. Sastry, \emph{A mathematical introduction to
  robotic manipulation}.\hskip 1em plus 0.5em minus 0.4em\relax CRC press,
  2017.

\bibitem{hartenberg1955kinematic}
R.~S. Hartenberg and J.~Denavit, ``A kinematic notation for lower pair
  mechanisms based on matrices,'' \emph{J. Appl. Mech.}, vol.~77, no.~2, pp.
  215--221, 1955.

\bibitem{libertiEuclideanDistanceGeometry2014}
L.~Liberti, C.~Lavor, N.~Maculan, and A.~Mucherino,
  ``\BIBforeignlanguage{en}{Euclidean {{Distance Geometry}} and
  {{Applications}}},'' \emph{\BIBforeignlanguage{en}{SIAM Rev.}}, vol.~56,
  no.~1, pp. 3--69, Jan. 2014.

\bibitem{Kingma2013-kw}
D.~P. Kingma and M.~Welling, ``Auto-encoding variational {B}ayes,'' in
  \emph{Intl. Conf. on Learning Representations (ICLR)}, Y.~Bengio and
  Y.~LeCun, Eds., 2014.

\bibitem{sohn_cvae_2015}
K.~Sohn, X.~Yan, and H.~Lee, ``Learning structured output representation using
  deep conditional generative models,'' in \emph{Advances in Neural Information
  Processing Systems - Volume 2 (NeurIPS}, 2015.

\bibitem{doersch2016tutorial}
C.~Doersch, ``Tutorial on variational autoencoders,'' \emph{arXiv preprint
  arXiv:1606.05908}, 2016.

\bibitem{satorras2021n}
V.~G. Satorras, E.~Hoogeboom, and M.~Welling, ``E (n) equivariant graph neural
  networks,'' in \emph{Intl. Conf. on Machine Learning (ICML)}, 2021.

\bibitem{elfwing2018sigmoid}
S.~Elfwing, E.~Uchibe, and K.~Doya, ``Sigmoid-weighted linear units for neural
  network function approximation in reinforcement learning,'' \emph{Neural
  Networks}, vol. 107, pp. 3--11, 2018.

\bibitem{ba2016layer}
J.~L. Ba, J.~R. Kiros, and G.~E. Hinton, ``Layer normalization,'' in
  \emph{Advances in Neural Information Processing Systems Deep Learning
  Symposium (NeurIPS)}, 2016.

\bibitem{LoshchilovH19}
I.~Loshchilov and F.~Hutter, ``Decoupled weight decay regularization,'' in
  \emph{Intl. Conf. on Learning Representations (ICLR)}, 2019.

\bibitem{schubert2017dbscan}
E.~Schubert, J.~Sander, M.~Ester, H.~P. Kriegel, and X.~Xu, ``Dbscan revisited,
  revisited: why and how you should (still) use dbscan,'' \emph{ACM
  Transactions on Database Systems (TODS)}, vol.~42, no.~3, pp. 1--21, 2017.

\bibitem{gilmer2017neural}
J.~Gilmer, S.~S. Schoenholz, P.~F. Riley, O.~Vinyals, and G.~E. Dahl, ``Neural
  message passing for quantum chemistry,'' in \emph{Intl. Conf. on Machine
  Learning (ICML)}, 2017.

\bibitem{velickovic2018graph}
P.~Velickovic, G.~Cucurull, A.~Casanova, A.~Romero, P.~Lio, and Y.~Bengio,
  ``Graph attention networks,'' in \emph{Intl. Conf. on Learning
  Representations (ICLR)}, 2018.

\bibitem{Kipf2016tc}
T.~N. Kipf and M.~Welling, ``{Semi-Supervised Classification with Graph
  Convolutional Networks},'' in \emph{Intl. Conf. on Learning Representations
  (ICLR)}, 2017.

\bibitem{hamilton2017inductive}
W.~Hamilton, Z.~Ying, and J.~Leskovec, ``Inductive representation learning on
  large graphs,'' \emph{Advances in Neural Information Processing Systems
  (NeurIPS)}, vol.~30, 2017.

\end{thebibliography}
